\documentclass[11pt]{article}

\usepackage[utf8]{inputenc}
\usepackage[T1]{fontenc}
\usepackage{arxiv}
\usepackage[numbers,sort&compress]{natbib}
\usepackage{graphicx}
\usepackage{adjustbox}
\usepackage{multirow}
\usepackage{amsmath,amssymb,amsfonts}
\usepackage{amsthm}
\usepackage{mathrsfs}
\usepackage{xcolor}
\usepackage{textcomp}
\usepackage{booktabs}
\usepackage{algorithm}
\usepackage{algorithmicx}
\usepackage{algpseudocode}
\usepackage{listings}
\usepackage{tabularx}
\usepackage{array}
\usepackage{microtype}
\usepackage[hidelinks]{hyperref}
\usepackage{placeins}
\usepackage{flafter}

\AddToHook{cmd/section/before}{\FloatBarrier}
\AddToHook{cmd/subsection/before}{\FloatBarrier}

\theoremstyle{definition}

\theoremstyle{remark}

\renewcommand{\headeright}{TongGuOCR}
\renewcommand{\shorttitle}{TongGuOCR}
\title{TongGuOCR: A Layout-Aware and Token-Augmented OCR MLLM for Chinese Historical Documents}

\author{
\textbf{Zhongheng Zhou}\textsuperscript{1,*} \quad
\textbf{Yi Sun}\textsuperscript{1,*} \quad
\textbf{Huiguo He}\textsuperscript{1,*,$\dagger$} \quad
\textbf{Yuyi Zhang}\textsuperscript{1} \quad
\textbf{Peirong Zhang}\textsuperscript{1} \\
\textbf{Yulin Fang}\textsuperscript{1} \quad
\textbf{Dezhi Peng}\textsuperscript{2} \quad
\textbf{Minghui Liao}\textsuperscript{2} \quad
\textbf{Lianwen Jin}\textsuperscript{1,$\dagger$} \\
\normalfont\footnotesize \textsuperscript{1}School of Electronic and Information Engineering, South China University of Technology, \\
\normalfont\footnotesize Guangzhou 510641, China \\
\normalfont\footnotesize \textsuperscript{2}Huawei Technologies Co., Ltd., Dongguan 511700, China \\
\normalfont\footnotesize \textsuperscript{*}Equal contribution. \quad
\textsuperscript{$\dagger$}Corresponding authors.
}

\date{}

\begin{document}

\maketitle

\begin{abstract}
Chinese historical documents preserve valuable cultural heritage, but many collections remain accessible only as scanned page images, preventing full-text retrieval, collation, and computational analysis. Optical character recognition (OCR) can bridge this gap, but accurate transcription remains challenging because historical documents often contain complex layouts, rare characters, and nontrivial reading orders. We propose \textbf{TongGuOCR}, a layout-aware and token-augmented multimodal large language model (MLLM) for OCR of Chinese historical documents. First, a Layout-Aware Preprocessing module constructs and refines locally coherent recognition blocks to preserve local context while reducing interference across regions. Second, a Token-Augmented Recognition module augments the transcription target at two complementary levels: character-level vocabulary expansion gives each rare glyph a direct one-token representation and shortens its decoding path, while line-to-line transition modeling injects discrete spatial displacement tokens that guide the decoder along complex reading paths without requiring precise coordinates. Experiments on two Chinese historical document OCR benchmarks show that TongGuOCR outperforms representative traditional task-specific OCR models, general-purpose MLLMs, and OCR-oriented MLLMs. On the more challenging M\textsuperscript{5}HisDoc benchmark,
TongGuOCR achieves 93.76 AR and reduces NED from 10.43 to 6.15 and RO-ED from 7.53 to 3.49 relative to the best competing score for each metric. An online demo is available at \url{https://jzzh2004.github.io/TongGuOCR}.
\end{abstract}

\section{Introduction}

Chinese historical documents preserve textual records spanning thousands of years and constitute a vital part of East Asian written cultural heritage~\cite{Wang2025-1,Guo2025,Li2026,Wang2026,lombardi2020survey,Sturgeon_2020}. To preserve and broaden access to these collections, 
large-scale digitization programs have converted many of them into scanned page images~\cite{sturgeon2019ctext}. However, page images alone cannot support full-text retrieval, textual collation, or the computational analysis increasingly required in digital humanities and heritage science~\cite{deweerdt2020markus}. Optical character recognition (OCR) bridges this gap by converting scanned pages into machine-readable text, enabling cross-document collation, entity extraction~\cite{Xu2025}, ancient-to-modern translation~\cite{Zhao2025}, knowledge graph construction~\cite{Liu2025}, and large-scale computational analysis of classical texts~\cite{cao2025tonggu}. OCR therefore serves as a foundational enabling technology for the scholarly and computational use of digitized historical document collections~\cite{sturgeon2018premodernocr}.

\begin{figure}[!htbp]
    \centering
    \includegraphics[width=\textwidth, trim=250 100 425 50, clip]{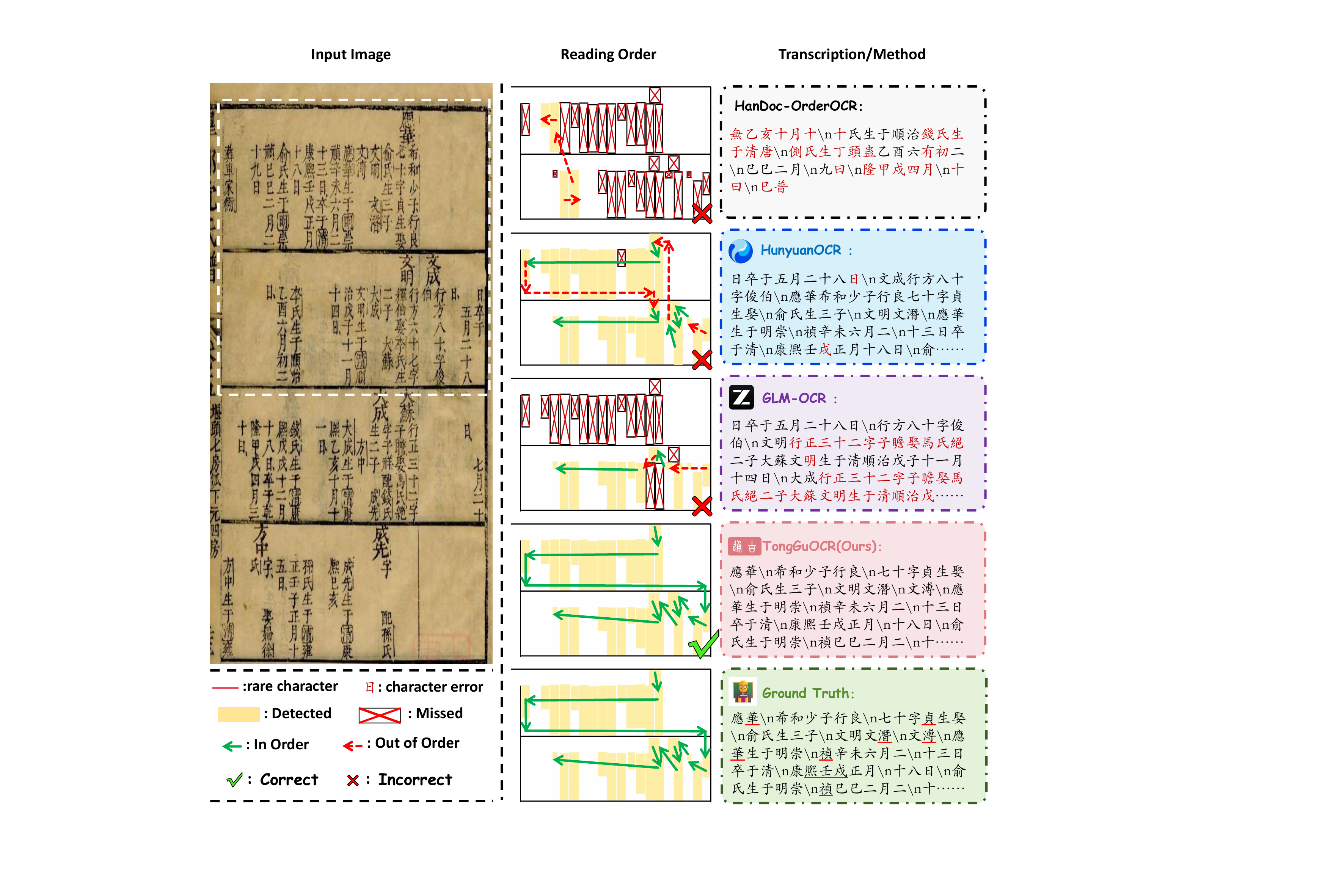}
    \caption{Recognition results illustrating the challenges of Chinese historical document OCR. The white dashed box in the input image marks the region used for the reading-order and transcription comparison. This example contains a table-like layout, a nontrivial reading order, and rare historical characters. We compare TongGuOCR with representative OCR systems, including HanDoc-OrderOCR~\cite{handoc_orderocr}, HunyuanOCR~\cite{hunyuanocr}, and GLM-OCR~\cite{glmocr}. Arrows visualize the reading paths aligned to the source region, while red characters indicate recognition errors or order-related mismatches.
    TongGuOCR follows the natural reading sequence of the page and accurately transcribes the selected content in this case.}
    \label{fig:intro_case}

\end{figure}

Earlier OCR studies on Chinese historical documents mostly adopted multi-stage pipelines that combine layout analysis, text-line or character detection, recognition, and reading-order recovery~\cite{sturgeon2018premodernocr,inner_engine,MTHv2,wang2019rarebookocr}. These methods provide explicit structural modeling and fine-grained control over page decomposition and recognition. More recently, general-purpose multimodal large language models (MLLMs) and OCR-oriented MLLMs have substantially advanced document OCR through large-scale visual-language pretraining, direct image-to-text generation, and integrated layout analysis, reading-order prediction, and recognition~\cite{qwen3vl,internvl3.5,hunyuanocr,deepseekocr2,glmocr,mineru25pro}.

However, these methods still face significant challenges when applied to Chinese historical documents.
First, page layouts are intricate and diverse, often mixing vertical columns, interlinear annotations, and irregular text blocks. Neither end-to-end OCR models nor traditional pipelines fully address this challenge. End-to-end models typically take full-page images as input, but resizing can reduce image resolution, and unrelated regions can interfere with recognition. Traditional pipelines instead detect and recognize individual text lines, preserving character detail but losing the local context needed for coherent transcription. A suitable intermediate recognition granularity is therefore needed to balance character visibility and contextual coverage.

Second, Chinese historical documents contain a large and long-tailed inventory of rare characters. Traditional classification-based recognizers rely on a predefined character set, leaving characters outside the label space unsupported and resulting in poor recognition of rare categories. MLLM-based recognizers face a different limitation: tokenizers trained primarily on modern text often decompose rare characters into multiple byte-level or subword fragments. This fragmentation weakens the correspondence between a visual glyph and its output representation, requiring multiple decoding steps for a single character.

Third, complex reading orders make line-to-line transitions difficult during transcription, even after layout analysis and region cropping. Multiple vertical columns, interlinear annotations, and marginal text can interrupt the regular reading flow, leaving the decoder uncertain about where to continue after completing a line. Without explicit spatial guidance, this ambiguity may lead to skipped or repeated lines and cross-line confusion. Supervision of the spatial displacement between successive lines is therefore needed to help the decoder follow the intended reading order. These challenges often occur simultaneously, leading to coupled errors in character recognition and reading-order recovery, as illustrated in Figure~\ref{fig:intro_case}.

To address these challenges, we propose \textbf{TongGuOCR}, a layout-aware and token-augmented OCR MLLM for Chinese historical documents. TongGuOCR consists of two modules: Layout-Aware Preprocessing and Token-Augmented Recognition. Specifically, \textbf{Layout-Aware Preprocessing} addresses the recognition-granularity problem by grouping consecutive text lines into locally coherent blocks and refining them into OCR-friendly crops. This process preserves local context while reducing interference from unrelated regions. \textbf{Token-Augmented Recognition} addresses the remaining two challenges through two complementary designs. Character-level vocabulary expansion gives rare characters single-token representations and thereby removes the fragmentation introduced by a tokenizer trained on modern text, while line-to-line transition modeling turns the implicit reading order into explicit supervision at each line boundary. A three-stage supervised fine-tuning (SFT) strategy progressively adapts the recognition model to historical documents and aligns its training inputs with the recognition blocks used during inference.

The main contributions of this work are as follows:
\begin{itemize}
    \item We propose \textbf{TongGuOCR}, a layout-aware and token-augmented OCR MLLM for Chinese historical documents. TongGuOCR integrates two complementary modules: Layout-Aware Preprocessing and Token-Augmented Recognition.
    
    \item We introduce a \textbf{Layout-Aware Preprocessing} module that constructs locally coherent recognition blocks from text-line boxes and refines them into OCR-friendly crops. These recognition blocks preserve character visibility while retaining sufficient local context.
    
    \item We introduce a \textbf{Token-Augmented Recognition} module with two complementary designs. Character-level vocabulary expansion improves the representation of rare characters, and line-to-line transition modeling provides spatial guidance during transcription.
    
    \item Experiments on two Chinese historical document OCR benchmarks show that TongGuOCR outperforms representative traditional task-specific OCR models, general-purpose MLLMs, and OCR-oriented MLLMs, achieving AR scores of 97.93 and 93.76 and RO-ED scores of 1.51 and 3.49 on MTHv2 and M\textsuperscript{5}HisDoc, respectively. Ablation studies verify the effectiveness of both modules and the three-stage SFT strategy.
\end{itemize}

\section{Related Work}
\label{sec:related_work}

\subsection{Traditional Historical Document OCR}

OCR for Chinese historical documents has traditionally been studied under a modular document-analysis paradigm. Instead of directly transcribing a full page in an end-to-end manner, earlier approaches usually decompose the task into stages such as image preprocessing, layout analysis, text-line or character localization, recognition, reading-order recovery, and post-correction~\cite{sturgeon2018premodernocr,inner_engine,MTHv2,SIHANG2020107503,wang2019rarebookocr}. Within this paradigm, prior studies have developed OCR systems for traditional Chinese texts~\cite{sturgeon2018premodernocr,inner_engine,wang2019rarebookocr}, investigated dense character detection and localization in historical document pages~\cite{8364534,SIHANG2020107503}, and jointly considered layout analysis, character detection, and recognition for historical document digitization~\cite{MTHv2}. Recent datasets and benchmarks, such as M\textsuperscript{5}HisDoc and HisDoc1B, further support research on multi-style and large-scale Chinese historical document analysis~\cite{shi2023m5hisdoc,hisdoc1b}. These works provide explicit structural modeling of document pages and enable fine-grained processing at the region, line, or character level.

Despite their explicit structural modeling, traditional historical-document OCR systems commonly treat individual text lines or characters as separate recognition inputs. This fine-grained formulation preserves local character resolution but provides limited context across neighboring lines. Moreover, traditional recognizers usually rely on predefined character sets, while page-level reading sequences are often reconstructed through separately designed geometric rules. These characteristics make rare characters and complex local line transitions difficult to handle. They motivate OCR approaches that retain explicit layout processing while using a more suitable recognition granularity, more flexible character representations, and explicit spatial guidance during transcription.

\subsection{Large Models for Document OCR}

Beyond traditional approaches, recent MLLMs have substantially advanced document OCR and document parsing. General-purpose MLLMs perform OCR through visual prompting and instruction following, benefiting from broad visual-language pretraining and semantic reasoning ability~\cite{qwen3vl,internvl3.5,qwen3.5,qwen37plus,gemini31pro,gpt55}. OCR-oriented MLLMs are designed specifically for document recognition and parsing. They include models that directly generate plain text, Markdown, or structured representations from document images~\cite{deepseekocr,hunyuanocr,dotsocr,fireredocr,deepseekocr2,dong2026qianfanocrunifiedendtoendmodel}, as well as MLLM-based systems that integrate layout analysis, recognition, reading-order prediction, and structure reconstruction~\cite{cui2026paddleocrvl15multitask09bvlm,mineru25pro,monkeyocr,mineru25,paddleocrvl16,glmocr,lggpt2025zhang}. These methods have shown strong performance on modern document images and complex document parsing tasks.

Nevertheless, these systems are primarily optimized for general document domains and do not explicitly account for characteristics of Chinese historical materials, such as complex layouts, historical characters, and nontrivial line transitions. This domain mismatch limits their reliability in faithful historical-document transcription and motivates task-specific adaptation.

\section{Methods}

\begin{figure}[!htbp]
    \centering
    \includegraphics[width=\textwidth, trim=230 210 230 210, clip]{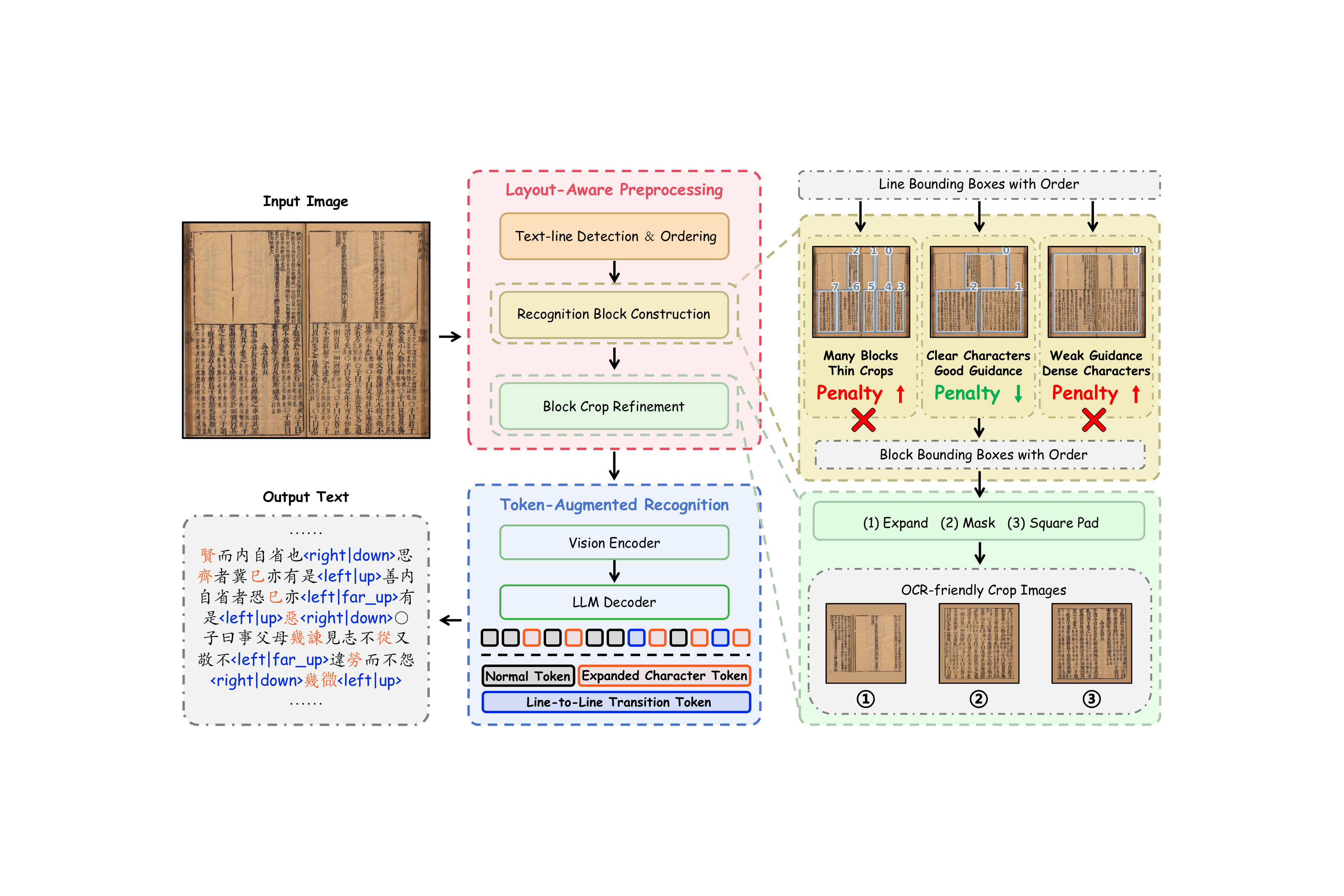}
    \caption{
    Overview of TongGuOCR. TongGuOCR consists of two modules. Layout-Aware Preprocessing obtains text-line boxes, constructs recognition blocks, and refines them into OCR-friendly crops. Token-Augmented Recognition then processes these crops using character-level vocabulary expansion and line-to-line transition modeling, after which the resulting transcriptions are merged into the page-level output. The enlarged panels on the right illustrate how block construction and crop refinement affect OCR inputs. Over-segmented candidates produce many narrow crops with limited context, while under-segmented candidates contain dense characters and provide weak layout guidance; both receive higher penalties. Balanced blocks preserve clearer characters and useful local layout context, resulting in a lower penalty and more OCR-friendly inputs.
    }
    \label{fig:overview}
\end{figure}

As illustrated in Figure~\ref{fig:overview}, TongGuOCR consists of two modules. Layout-Aware Preprocessing (Section~\ref{sec:layout_preprocessing}) converts a full page into an ordered sequence of OCR-friendly recognition blocks that preserve local context while reducing interference from unrelated regions. Token-Augmented Recognition (Section~\ref{sec:token_augmented_ocr}) then transcribes each block, improving rare-character representation and providing spatial guidance during decoding. TongGuOCR employs a three-stage supervised fine-tuning strategy to adapt the recognition model to historical documents and align its training inputs with the recognition-block crops used during inference (Section~\ref{sec:training_strategy}).

\subsection{Layout-Aware Preprocessing}
\label{sec:layout_preprocessing}

A key design choice in TongGuOCR is the recognition granularity. Whole-page input requires aggressive resizing that blurs small characters~\cite{UReader}, whereas line-level recognition yields elongated crops, loses neighboring context, and needs one forward pass per line~\cite{TextMonkey}. Recognition blocks offer an intermediate granularity, but grouping lines is nontrivial, since visual adjacency need not match reading-order adjacency (e.g., interlinear annotations) and the ideal block size is layout-dependent. We therefore restrict each block to consecutive lines in the reading order and organize the grouping into three steps, detailed below: Candidate Block Construction, Optimal Segmentation Selection, and Block Crop Refinement.

\subsubsection{Candidate Block Construction}

\noindent\textbf{Text-line boxes and reading order.}
We first obtain the text-line boxes of a page together with their reading order, denoted \(L=[l_1,l_2,\ldots,l_n]\). During training, the boxes and their order are taken from annotations; during inference, they are produced by an external historical-document layout module~\cite{inner_engine}. This order determines which lines may be grouped consecutively and is preserved throughout the resulting block sequence.

\noindent\textbf{Candidate blocks and segmentations.}
For a start index \(i\) and an exclusive end boundary \(j\), we define a candidate recognition block as
\begin{equation}
B_{i,j}=(l_i,l_{i+1},\ldots,l_{j-1}),
\qquad 1\leq i<j\leq n+1.
\end{equation}
For example, \(B_{3,6}=(l_3,l_4,l_5)\). Restricting each candidate to consecutive lines preserves the reading order in \(L\) and avoids grouping spatially adjacent lines that are discontinuous in the sequence. A valid segmentation \(S=[B_1,\ldots,B_K]\) selects non-overlapping candidate blocks that jointly cover \(L\), so that every text line is assigned to exactly one block without changing its order. We denote the set of all such segmentations by \(\mathcal{S}(L)\).

\subsubsection{Optimal Segmentation Selection}

\noindent\textbf{Penalty-based segmentation objective.}
Let \(\mathcal{P}(S)\) denote the total layout penalty of a complete segmentation \(S\). We select the valid segmentation with the lowest total penalty:
\begin{equation}
S^*=\arg\min_{S\in\mathcal{S}(L)}\mathcal{P}(S),
\end{equation}
where
\begin{equation}
\mathcal{P}(S)
=
\lambda_{\mathrm{blk}}\sum_{t=1}^{K}p_{\mathrm{blk}}(B_t)
+
\lambda_{\mathrm{adj}}\sum_{t=1}^{K-1}p_{\mathrm{adj}}(B_t,B_{t+1})
+
\lambda_{\mathrm{page}}p_{\mathrm{page}}(K).
\end{equation}
The three penalty groups evaluate complementary aspects of a segmentation. The block-level penalty \(p_{\mathrm{blk}}\) measures whether each candidate balances character visibility and local layout context, corresponding to the different granularities illustrated in Figure~\ref{fig:overview}. The adjacent-block penalty \(p_{\mathrm{adj}}\) discourages geometrically unstable boundaries, such as severe overlap or containment between successive blocks. The page-level penalty \(p_{\mathrm{page}}\) regularizes the total number of blocks to avoid both whole-page grouping and excessive fragmentation. The non-negative coefficients \(\lambda_{\mathrm{blk}}\), \(\lambda_{\mathrm{adj}}\), and \(\lambda_{\mathrm{page}}\) control their relative contributions.

\noindent\textbf{Dynamic programming solution.}
The consecutive-interval representation gives the segmentation problem an ordered optimal substructure: a partial segmentation ending at boundary \(j\) can be formed by appending a candidate \(B_{i,j}\) to a predecessor that ends at \(i\). We therefore use dynamic programming over interval boundaries and block counts to obtain the minimum-penalty segmentation within the retained candidate space. After backtracking, a lightweight local merge step absorbs isolated tiny blocks into geometrically compatible neighboring blocks to reduce unstable fragmentation. Detailed geometric cues, candidate pruning, the dynamic-programming recurrence, backtracking, and local merging are provided in supplementary material.

\subsubsection{Block Crop Refinement}

Each selected block is finally converted into an OCR-friendly crop with reduced cross-block interference and more stable geometry, as shown in the lower-right panel of Figure~\ref{fig:overview}. Existing OCR systems commonly resize document regions using high-resolution or dynamic-resolution visual encoders~\cite{Extending,Large,paddleocrvl} and regularize them into square or fixed-size inputs~\cite{UNIT}. Consequently, tightly bounded or highly elongated crops may lose boundary context or undergo excessive downsampling. We therefore expand each selected block box by a small margin, mask overlapping text lines belonging to other blocks, and place the processed crop on a padded square canvas while preserving its aspect ratio.

\noindent\textbf{Boundary expansion.}
We first enlarge each selected block box by a small margin, so that characters near the block boundary retain sufficient surrounding context after cropping.

\noindent\textbf{Non-target line masking.}
For block \(B_t\), let \(M_t^{\mathrm{own}}\) denote the mask of its own text lines and \(M_t^{\mathrm{nt}}\) the mask of overlapping non-target lines. The region to be removed is
\begin{equation}
M_t=M_t^{\mathrm{nt}}\setminus M_t^{\mathrm{own}}.
\end{equation}
Masked pixels are filled with a locally estimated background color.

\noindent\textbf{Aspect-ratio-preserving padding.}
If the resulting crop has width \(w_t\) and height \(h_t\), it is centered on a square canvas with side length
\begin{equation}
s_t=\alpha\max(w_t,h_t),
\end{equation}
where \(\alpha\) is a fixed padding scale. The refined crops preserve their original aspect ratios and block order before being passed to the recognition module. Further implementation details are provided in the supplementary material.

\subsection{Token-Augmented Recognition}
\label{sec:token_augmented_ocr}

\begin{figure}[!htbp]
\centering
\includegraphics[width=0.9\textwidth,trim=80 50 60 85,
    clip]{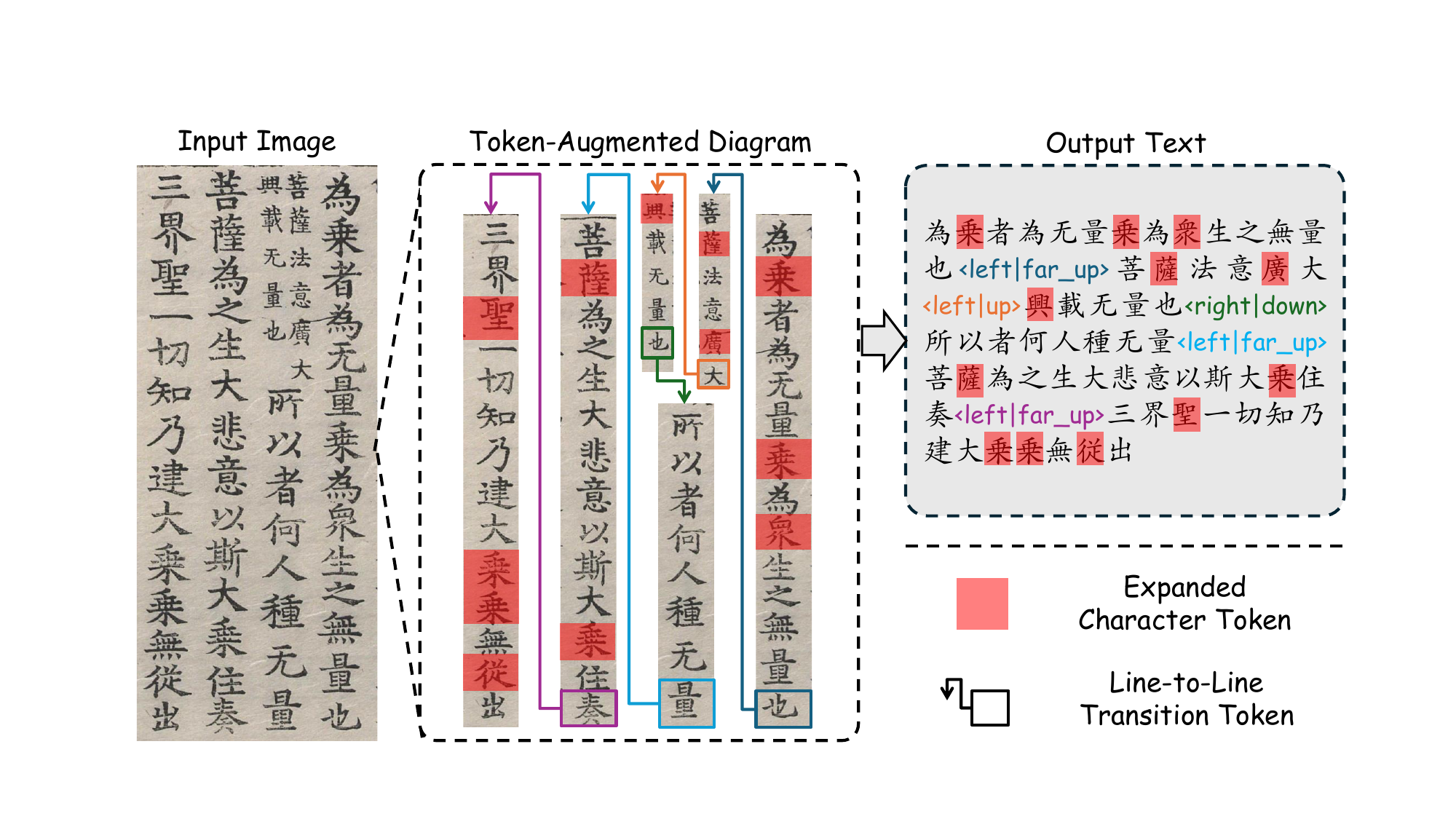}
\caption{
Illustration of the Token-Augmented Recognition target construction. The transcription is represented as an augmented sequence containing ordinary text tokens, expanded character tokens, and line-to-line transition tokens. The highlighted Chinese characters indicate previously fragmented characters that are represented as single tokens after vocabulary expansion. Transition tokens, such as \texttt{<left|far\_up>} and \texttt{<right|down>}, encode the coarse spatial displacement from the end of one text line to the beginning of the next.
}
\label{fig:token_augmented_ocr}
\end{figure}

The recognition module produces a token sequence rather than isolated character predictions. Therefore, the construction of the transcription target affects both character generation and the spatial organization of the output sequence. We augment the target sequence at two complementary levels. First, we expand the tokenizer with single Chinese character tokens so that rare characters can be represented more compactly. Second, we insert line-to-line transition tokens between adjacent line transcriptions. These tokens encode the coarse spatial displacement from the end of one line to the beginning of the next, providing auxiliary spatial cues during autoregressive decoding. Figure~\ref{fig:token_augmented_ocr} illustrates the resulting augmented sequence construction.

\subsubsection{Character-Level Vocabulary Expansion}
\label{sec:vocab_expansion}

Faithful transcription of Chinese historical documents requires character-level output. However, tokenizers inherited from pretrained MLLMs are usually optimized for modern text and often lack standalone tokens for rare Chinese characters. A single rare character may therefore be decomposed into multiple byte-level or subword tokens, requiring several autoregressive decoding steps to transcribe one visual character instance. Representing such a character as a single token makes its visual-to-output correspondence more direct and shortens its decoding path, potentially reducing character-level generation errors and reliance on contextual language priors.

\noindent\textbf{Candidate character mining.}
Let $\mathcal{V}_0$ denote the vocabulary of the original tokenizer, and let $\mathrm{Tok}_0(\cdot)$ denote its tokenization function. We train an auxiliary SentencePiece tokenizer with the byte-pair encoding (BPE) model~\cite{kudo-richardson-2018-sentencepiece} on the training corpus. The validation and test splits are excluded. This auxiliary tokenizer is used only to mine candidate characters, rather than to replace the original tokenizer. We remove special tokens such as \texttt{<unk>}, \texttt{<s>}, and \texttt{</s>}, strip the SentencePiece whitespace marker when present, and retain candidate characters from the Unicode CJK Unified Ideographs, CJK Compatibility Ideographs, and related extension blocks.\footnote{CJK stands for Chinese, Japanese, and Korean. For Unicode's treatment of Han ideographs across these writing systems, see \url{https://www.unicode.org/faq/han_cjk.html}.} Let $\mathcal{C}_{\mathrm{CJK}}$ denote this candidate character set.

\noindent\textbf{Character selection criterion.}
From these candidates, we add only characters satisfying the following condition:
\begin{equation}
\mathcal{A}
=
\{c \mid |c|=1,\;
c\in\mathcal{C}_{\mathrm{CJK}},\;
c\notin\mathcal{V}_0,\;
|\mathrm{Tok}_0(c)|>1\}.
\end{equation}
That is, a new character token is added only when it is a single character, is absent from the original vocabulary, and would otherwise be fragmented into multiple tokens by the original tokenizer. After vocabulary expansion, the model embedding matrix is resized using the default model resizing routine, and the newly added embeddings are optimized during training.

\subsubsection{Line-to-Line Transition Modeling}
\label{sec:line_transition}

Layout analysis and region cropping reduce page-level complexity but do not eliminate ambiguity in transitions between text lines, since both full pages and cropped regions may contain multiple columns, interlinear annotations, or other irregular arrangements. After transcribing one line, the model must locate the visual region of the next line, and errors here cause skipped or repeated lines and cross-line confusion. To supervise this process, we insert a discrete transition token between adjacent line transcriptions that encodes the coarse spatial displacement from the end of one line to the beginning of the next, guiding the decoder toward the next relevant region without requiring precise coordinates.

\noindent\textbf{Transition token insertion.}
Given a dataset-annotated sequence of text lines \(L=[l_1,l_2,\ldots,l_n]\), we augment the transcription target by inserting a line-to-line transition token between each pair of adjacent line transcriptions:
\begin{equation}
Y_{\mathrm{trans}}
= T(l_1) \oplus r_1 \oplus T(l_2) \oplus \cdots
\oplus r_{n-1} \oplus T(l_n),
\end{equation}
where \(T(\cdot)\) denotes tokenization with the expanded tokenizer, \(\oplus\) denotes sequence concatenation, and \(r_i\) encodes the coarse spatial displacement from the end of \(l_i\) to the beginning of \(l_{i+1}\).

\noindent\textbf{Line-to-line displacement computation.}
For each adjacent pair of text lines, $r_i$ is determined by the displacement from the end anchor of the current line to the start anchor of the next line:
\begin{equation}
\Delta_i =
\left(
 x_{i+1}^{\mathrm{start}} - x_i^{\mathrm{end}},
 y_{i+1}^{\mathrm{start}} - y_i^{\mathrm{end}}
\right).
\end{equation}
The start and end anchors are defined as the centers of the first and last dataset-annotated characters in the corresponding line, respectively. This definition aligns the transition token with the actual spatial movement between successive line transcriptions. The horizontal and vertical offsets are normalized by the image width and height. Under the image coordinate system, the sign of the horizontal offset determines \texttt{left} or \texttt{right}, and the sign of the vertical offset determines \texttt{up} or \texttt{down}.

\noindent\textbf{Coarse displacement quantization.}
To convert continuous line-to-line displacements into discrete transition tokens, we apply a fixed coarse quantization rule. The goal of this quantization is not to model precise geometry, but to provide a compact set of stable spatial categories for auxiliary supervision at line boundaries. Let $\delta$ denote a near-zero threshold and $\tau$ denote a long-range threshold. For each axis, offsets whose magnitudes are smaller than $\delta$ are omitted, whereas offsets whose magnitudes exceed $\tau$ are assigned the prefix \texttt{far\_}. If both horizontal and vertical offsets are negligible, we retain the axis with the larger normalized displacement to ensure that each adjacent line pair receives a valid transition token. The horizontal and vertical label sets are
\begin{equation}
\mathcal{R}_x =
\{
\texttt{left},
\texttt{right},
\texttt{far\_left},
\texttt{far\_right}
\},
\end{equation}
and
\begin{equation}
\mathcal{R}_y =
\{
\texttt{up},
\texttt{down},
\texttt{far\_up},
\texttt{far\_down}
\}.
\end{equation}

\noindent\textbf{Transition vocabulary construction.}
The final transition token vocabulary is constructed as
\begin{equation}
\begin{aligned}
\mathcal{R}
={}& \{\langle x\rangle \mid x \in \mathcal{R}_x\} \\
&\cup \{\langle y\rangle \mid y \in \mathcal{R}_y\} \\
&\cup \{\langle x|y\rangle \mid x \in \mathcal{R}_x,\;
y \in \mathcal{R}_y\}.
\end{aligned}
\end{equation}
This construction yields a compact set of 24 possible transition tokens. For instance, the transition from the bottom of one vertical column to the top of the neighboring column can be represented by \texttt{<left|far\_up>}, while a short movement into an annotation region or a return to the main text stream can be represented by tokens such as \texttt{<right|down>} or \texttt{<left|up>}. Thus, each token describes the coarse relative location of the next line.

By predicting transition tokens jointly with text, the decoder learns to associate each line boundary with the approximate location of the next text line. These tokens therefore provide auxiliary spatial supervision during transcription, supporting more stable and accurate output generation.

\subsection{Training Strategy}
\label{sec:training_strategy}

TongGuOCR is optimized through three successive SFT stages. Stage 1 performs coarse domain adaptation with diverse page-level samples and develops broad familiarity with Chinese historical documents. Stage 2 refines page-level recognition with human-annotated data. Stage 3 uses recognition-block crops generated by Layout-Aware Preprocessing, thereby aligning the training and inference inputs. All three stages use the same augmented target representation and autoregressive objective.

Given an input image \(I\) and an augmented transcription sequence \(Y=(y_1,y_2,\ldots,y_T)\), the model is optimized with the standard autoregressive language modeling objective:
\begin{equation}
\mathcal{L}_{\mathrm{LM}}
=
-\frac{1}{T}
\sum_{t=1}^{T}
\log p_{\theta}(y_t \mid I,  y_{<t}).
\end{equation}
The target sequence \(Y\) contains ordinary text tokens and line-to-line transition tokens; text generation and transition-token prediction are therefore optimized jointly under the same next-token prediction objective.

\section{Experiments}

\subsection{Datasets}
\label{sec:datasets}
\begin{table}[!htbp]
\centering
\caption{Dataset annotation source and split usage. SFT, Select, and Eval denote supervised fine-tuning, checkpoint selection, and final evaluation, respectively.}
\label{tab:datasets}
\small
\setlength{\tabcolsep}{9pt}
\renewcommand{\arraystretch}{1.12}
\begin{tabular}{l l c c c}
\toprule
Dataset & Annotation source & Train & Val & Test \\
\midrule
HisDoc1B & Semi-automatic & 3.16M / SFT & -- & -- \\
MTHv2 & Human & 2,399 / SFT & -- & 800 / Eval \\
M\textsuperscript{5}HisDoc & Human & 2,000 / SFT & 1,000 / Select & 1,000 / Eval \\
\bottomrule
\end{tabular}
\end{table}
We conduct experiments on three publicly released Chinese historical document datasets: HisDoc1B~\cite{hisdoc1b}, MTHv2~\cite{MTHv2}, and M\textsuperscript{5}HisDoc~\cite{shi2023m5hisdoc}. HisDoc1B serves as a large-scale dataset for SFT, while M\textsuperscript{5}HisDoc and MTHv2 are used for SFT and evaluation. Table~\ref{tab:datasets} summarizes the annotation source and split usage of each dataset.

\noindent\textbf{HisDoc1B.} This large-scale Chinese historical document dataset contains 40,281 books, 3,163,330 images, 1,082,544,808 characters, and 30,615 character categories. It was constructed through a semi-automatic annotation pipeline designed for large-scale historical document collection and annotation. In our experiments, HisDoc1B serves as a large-scale training set that exposes the model to diverse historical document styles, page layouts, character forms, and transcription lengths.

\noindent\textbf{MTHv2.} Built from the Tripitaka Koreana in Han and Multiple Tripitaka in Han collections, this dataset contains 3,199 document images, 1,081,678 characters, and 6,733 character categories. In our experiments, we follow the official train/test split, which consists of 2,399 training images and 800 testing images. MTHv2 does not provide a validation split.

\noindent\textbf{M\textsuperscript{5}HisDoc.} This multi-style Chinese historical document dataset contains 4,000 images, 403,824 text lines, 4,367,392 characters, and 16,151 character categories. It covers diverse layouts, document types, calligraphic styles, and backgrounds. In our experiments, we follow the regular setting and use the official split, with 2,000 images for training, 1,000 images for validation, and 1,000 images for evaluation.

Note that HisDoc1B does not include any samples from the test sets of MTHv2 or M\textsuperscript{5}HisDoc.

\subsection{Evaluation Metrics}
\label{sec:evaluation_metrics}
The OCR-related metrics are implemented using a modified version of the evaluation protocol from MCS-Bench~\cite{liu-etal-2025-mcs}. Unlike MCS-Bench, we do not convert model outputs to Simplified Chinese before evaluation. For each metric, scores are computed on plain textual transcriptions for individual samples, averaged across all evaluation samples, and reported as percentages.

\noindent\textbf{Accuracy and correct rate (AR and CR).} We use AR and CR as the primary OCR accuracy metrics. For a given sample, let $G$ denote the reference transcription and $P$ denote the predicted transcription. Based on the Levenshtein alignment between $G$ and $P$, let $S$, $D$, and $I$ be the numbers of substitutions, deletions, and insertions, respectively. Let $N=|G|$ be the number of characters in the reference transcription. AR and CR are defined as
\begin{equation}
\mathrm{AR} = \frac{N-S-D-I}{N},
\end{equation}
and
\begin{equation}
\mathrm{CR} = \frac{N-S-D}{N}.
\end{equation}
AR penalizes substitutions ($S$), deletions ($D$), and insertions ($I$), whereas CR penalizes only $S$ and $D$ and is therefore unaffected by insertion errors. Notably, AR can take negative values: $\mathrm{AR}<0$ when $S+D+I>N$, meaning that the total number of substitution, deletion, and insertion errors exceeds the number of ground-truth characters.

\noindent\textbf{Normalized edit distance (NED).} We report NED:
\begin{equation}
\mathrm{NED} =
\frac{\mathrm{EditDistance}(P,G)}
{\max(|P|,|G|)},
\end{equation}
where $\mathrm{EditDistance}(\cdot,\cdot)$ denotes the Levenshtein edit distance. Lower NED indicates better sequence-level agreement.

\noindent\textbf{Character-inventory precision, recall, and F1.} We report character-inventory precision, recall, and F1 score as auxiliary coverage metrics. Let $C_P$ and $C_G$ denote the sets of unique characters appearing in the prediction and reference, respectively. They are defined as
\begin{equation}
\mathrm{Precision} = \frac{|C_P \cap C_G|}{|C_P|},
\end{equation}
\begin{equation}
\mathrm{Recall} = \frac{|C_P \cap C_G|}{|C_G|},
\end{equation}
and
\begin{equation}
\mathrm{F1} =
\frac{2 \cdot \mathrm{Precision} \cdot \mathrm{Recall}}
{\mathrm{Precision}+\mathrm{Recall}}.
\end{equation}
These set-based metrics measure character-inventory coverage, but they do not account for character order or repeated occurrences. Thus, F1 reflects the overlap between the predicted and reference character inventories, whereas AR evaluates occurrence-level accuracy over the full transcription sequence by penalizing substitutions, deletions, and insertions.

\noindent\textbf{Character-level BLEU.} Character-level BLEU serves as an auxiliary sequence-similarity metric. Both the reference and prediction are segmented into character sequences, and sentence-level BLEU is computed with smoothing.

\noindent\textbf{Reading-order edit distance (RO-ED).} To explicitly evaluate reading-order preservation, we follow the reading-order evaluation protocol of OmniDocBench~\cite{ouyang2024omnidocbenchbenchmarkingdiversepdf}, which measures reading order using normalized edit distance over matched textual components, and report a line-level RO-ED. RO-ED is computed on sequences of matched text-line indices. For each page, the ground-truth transcription is decomposed into an ordered list of text lines, and each line is assigned a reading-order index, forming the reference sequence $\mathbf{o}^{gt}=[1,2,\ldots,n]$. The predicted text lines are first matched to ground-truth lines according to textual similarity. The ground-truth indices of the matched lines are then sorted according to their positions in the model output, yielding the predicted order sequence $\mathbf{o}^{pred}=[\pi_1,\pi_2,\ldots,\pi_m]$, where $\pi_j$ denotes the ground-truth reading-order index of the $j$-th matched predicted line.

RO-ED is computed as the normalized edit distance between the two index sequences:
\begin{equation}
\mathrm{RO\text{-}ED}
=
\frac{
\mathrm{ED}_{seq}(\mathbf{o}^{gt}, \mathbf{o}^{pred})
}{
\max(|\mathbf{o}^{gt}|, |\mathbf{o}^{pred}|)
},
\end{equation}
where $\mathrm{ED}_{seq}$ denotes Levenshtein edit distance over discrete line indices. Lower RO-ED indicates better reading-order preservation, with 0 corresponding to a perfectly ordered prediction. We report the average RO-ED over all test pages.

Higher values are better for AR, CR, precision, recall, F1, and BLEU, whereas lower values are better for NED and RO-ED.

\subsection{Implementation Details}

\begin{table}[!htbp]
\centering
\caption{Implementation settings of the three-stage training strategy. MTH denotes MTHv2, M5H denotes M\textsuperscript{5}HisDoc, and H1B denotes HisDoc1B.}
\label{tab:stage_training}
\small
\setlength{\tabcolsep}{5pt}
\renewcommand{\arraystretch}{1.1}
\newcolumntype{Y}{>{\raggedright\arraybackslash}X}

\begin{tabularx}{0.9\textwidth}{@{}YYYY@{}}
\toprule
\textbf{Setting}
& \textbf{Stage 1}
& \textbf{Stage 2}
& \textbf{Stage 3} \\
\midrule

Training data
& H1B, MTH, M5H
& MTH, M5H
& MTH, M5H \\

Input granularity
& Page
& Page
& Block \\

Max image resolution
& \(768\times768\)
& \(1024\times1024\)
& \(1024\times1024\) \\

Max sequence tokens
& 6144
& 8192
& 8192 \\

Learning rate
& \(3\times10^{-5}\)
& \(1\times10^{-5}\)
& \(1\times10^{-5}\) \\

Epochs
& 3
& 10
& 15 \\

\bottomrule
\end{tabularx}
\end{table}
\noindent\textbf{Runtime environment.} Experiments were conducted under Python 3.11.13 and PyTorch 2.7.1 with Ascend NPU support, implemented with the LLaMA-Factory~\cite{zheng-etal-2024-llamafactory}, and executed on a server equipped with eight Huawei Ascend 910B2 NPUs, each with 64 GB HBM.

\noindent\textbf{Model configuration.} We adopt GLM-OCR~\cite{glmocr} as our foundation model. We first extend the original tokenizer with both the newly added single-character Chinese tokens and the line-to-line transition tokens, and then expand the model token embeddings accordingly. The original token embeddings are inherited from the pretrained model, while the newly added embedding rows are initialized with the default strategy. For line-to-line transition quantization, we use fixed heuristic thresholds of \(\delta=0.01\) and \(\tau=0.5\) after normalization. Here, \(\delta\) suppresses near-zero displacements caused by minor localization variations, while \(\tau\) marks an axis-wise movement spanning more than half of the page as a long-range transition. These thresholds are used throughout all experiments.

For Layout-Aware Preprocessing, we set \(\lambda_{\mathrm{blk}}=2\), \(\lambda_{\mathrm{adj}}=2\), and \(\lambda_{\mathrm{page}}=1\) throughout all experiments.

\noindent\textbf{Training configuration.} We adopt full-parameter SFT in all stages; that is, the vision encoder, multimodal projector, language model, and newly added token embeddings are all updated. All training stages share the same optimization configuration. We use the AdamW optimizer~\cite{loshchilov2017decoupled}, an adaptive optimizer with \(\beta_1=0.9\), \(\beta_2=0.999\), and \(\epsilon=10^{-8}\). Training follows a cosine decay schedule with a warmup ratio of 0.1, and gradients are clipped with a maximum norm of 1.0. Scaled dot-product attention is used as the attention backend. All training stages were conducted using bfloat16 (bf16) precision.

The stage-specific training configurations are summarized in Table~\ref{tab:stage_training}.

\textit{Stage 1: Coarse domain adaptation.} Stage 1 performs coarse domain adaptation using page-level training samples from HisDoc1B, MTHv2, and M\textsuperscript{5}HisDoc. The full HisDoc1B training set is used without subsampling. This stage adopts a maximum image resolution of \(768\times768\), a maximum sequence length of 6144 tokens, a learning rate of \(3\times10^{-5}\), and 3 training epochs.

\textit{Stage 2: High-quality page-level refinement.} Stage 2 performs high-quality page-level refinement using the human-annotated training splits of MTHv2 and M\textsuperscript{5}HisDoc. It adopts a maximum image resolution of \(1024\times1024\), a maximum sequence length of 8192 tokens, a learning rate of \(1\times10^{-5}\), and 10 training epochs.

\textit{Stage 3: Block-level inference alignment.} Stage 3 performs block-level inference alignment using recognition-block samples generated from the MTHv2 and M\textsuperscript{5}HisDoc training splits. These samples are produced by the same Layout-Aware Preprocessing pipeline employed during inference. This stage uses the same maximum image resolution, sequence length, and learning rate as Stage 2, and is trained for 15 epochs.

\noindent\textbf{Checkpoint selection.} Checkpoint selection is based on performance on the official validation split of M\textsuperscript{5}HisDoc. The final Stage 1 checkpoint at epoch 3 is used to initialize Stage 2, the validation-selected Stage 2 checkpoint at epoch 5 is used to initialize Stage 3, and the validation-selected Stage 3 checkpoint at epoch 3 is used for the main experimental results.

\noindent\textbf{Inference procedure.} During inference, the full page is first processed by Layout-Aware Preprocessing, which produces an ordered sequence of OCR-friendly crops. Each crop is then recognized autoregressively. Line-to-line transition tokens in each decoded sequence are replaced with newline characters to preserve line boundaries, and the resulting text transcriptions are concatenated in the sequence provided by Layout-Aware Preprocessing to obtain the final page-level transcription.

\subsection{Main Results}
\begin{table}[!htbp]
\centering
\scriptsize
\setlength{\tabcolsep}{4pt}
\renewcommand{\arraystretch}{1.05}
\caption{Performance comparison on MTHv2. AR, CR, NED, F1, P, R, BLEU, and RO-ED denote accuracy rate, correct rate, normalized edit distance, F1 score, precision, recall, BLEU, and reading-order edit distance, respectively. A dash denotes a result that was not reported. Higher values are better for AR, CR, F1, P, R, and BLEU, while lower values are better for NED and RO-ED. Bold and underlined values denote the best and second-best results among all compared methods, respectively. $^{\ast}$ indicates results reported under the text-line recognition setting and taken from ZCTRN~\cite{ZCTRN}; in this setting, input line images are constructed from the dataset-provided text-line annotations. $^\dagger$ indicates results reported in the MTHv2 paper~\cite{MTHv2}.}
\label{tab:main_mthv2}
\resizebox{\textwidth}{!}{%
\begin{tabular}{llcccccccc}
\toprule
\multirow{2}{*}{\textbf{Method}}
& \multirow{2}{*}{\textbf{Venue / Source}}
& \multicolumn{8}{c}{\textbf{MTHv2}} \\
\cmidrule(lr){3-10}
& & \textbf{AR}$\uparrow$ & \textbf{CR}$\uparrow$ & \textbf{NED}$\downarrow$ & \textbf{F1}$\uparrow$ & \textbf{P}$\uparrow$ & \textbf{R}$\uparrow$ & \textbf{BLEU}$\uparrow$ & \textbf{RO-ED}$\downarrow$ \\
\midrule
\multicolumn{10}{l}{\textit{Traditional task-specific OCR models}} \\
RAN~\cite{RAN}$^{\ast}$  &  PR 2020 & 91.56 & 91.79 & - & - & - & - & - & -\\
CRNN~\cite{CRNN}$^{\ast}$ & T-PAMI 2016  & 96.94 &97.15 & - & - & - & - & - & - \\
ZCTRN~\cite{ZCTRN}$^{\ast}$ &  ICDAR 2021 & \underline{97.42} & 97.62 & - & - & - & - & - & - \\
Ma et al.~\cite{MTHv2}$^\dagger$ &  ICFHR 2020  & 95.52 & 96.07  & - & - & - & - & - & - \\
HanDoc-OrderOCR~\cite{handoc_orderocr} & AAAI 2024 & 95.27 & 95.84 & 4.72 & 96.72 & 97.24 & 96.28 & 92.22 & 4.71 \\
Liu et al.~\cite{inner_engine}          &  NSR 2023     & 96.30 & 97.18 & 3.61 & 96.94 & 96.42 & 97.52 & 93.99 & 3.19 \\
\midrule
\multicolumn{10}{l}{\textit{General-purpose MLLMs}} \\
Qwen3-VL-2B~\cite{qwen3vl}          & arXiv 2025     & -376.72 & 71.61 & 54.71 & 76.94 & 88.51 & 72.37 & 40.12 & 32.43 \\
Qwen3-VL-8B~\cite{qwen3vl}          & arXiv 2025     & -29.34 & 82.70 & 33.38 & 85.43 & 90.51 & 83.27 & 61.33 & 21.06 \\
Qwen3.5-0.8B~\cite{qwen3.5}           & arXiv 2026 &  -69.63 & 70.59 & 38.38 & 79.24 & 86.09 & 75.39 & 53.38 & 31.99\\
Qwen3.5-9B~\cite{qwen3.5}           & arXiv 2026 &  62.17 & 80.74 & 23.05 & 86.63 & 89.51 & 84.68 & 69.68 & 18.97 \\
InternVL3.5-2B~\cite{internvl3.5}   & arXiv 2025     & -126.62 & 59.13 & 51.70 & 74.48 & 91.90 & 66.25 & 44.66 & 44.89\\
InternVL3.5-8B~\cite{internvl3.5}   & arXiv 2025     & -38.99 & 72.65 & 33.90 & 81.22 & 92.13 & 74.53 & 57.53 & 29.88\\
Qwen3.7-Plus~\cite{qwen37plus}      & Official API & 81.24 & 88.06 & 15.73 & 88.76 & 88.30 & 89.59 & 77.48 & 15.95 \\
MiniMax M3~\cite{minimaxm3}      & Official API & 67.54 & 75.15 & 27.69 & 81.65 & 85.47 & 79.40 & 63.48 & 26.01 \\
Kimi K2.6~\cite{kimik26}      & Official API & 86.40 & 90.66 & 11.79 & 92.14 & 92.40 & 92.10 & 81.97 & 10.57 \\
GPT-5.5~\cite{gpt55}      & Official API & 70.18 & 74.38 & 28.59 & 80.61 & 85.30 & 78.00 & 64.26 & 28.03 \\
Gemini 3.1 Pro~\cite{gemini31pro}      & Official API & 78.09 & 84.44 & 19.08 & 87.79 & 88.66 & 87.83 & 75.33 & 19.77 \\
\midrule
\multicolumn{10}{l}{\textit{OCR-oriented MLLMs}} \\
DeepSeek-OCR 2~\cite{deepseekocr2}  & arXiv 2026     & -540.58 & 26.55 & 83.83 & 41.69 & 71.87 & 33.66 & 13.56 & 68.81 \\
DeepSeek-OCR~\cite{deepseekocr}     & arXiv 2025     & -494.24 & 63.41 & 60.37 & 73.25 & 86.37 & 66.46 & 35.81 & 44.13 \\
FireRed-OCR~\cite{fireredocr}       & arXiv 2026     & -198.23 & 82.38 & 26.66 & 85.78 & 91.40 & 83.04 & 66.73 & 17.24 \\
dots.ocr~\cite{dotsocr}             & arXiv 2025     &   -3.27 & 84.11 & 23.65 & 87.71 & 90.84 & 86.02 & 69.09 & 16.43 \\
Qianfan-OCR~\cite{dong2026qianfanocrunifiedendtoendmodel} & arXiv 2026 & 62.79 & 86.90 & 18.23 & 88.35 & 88.86 & 88.48 & 73.67 & 16.42 \\
HunyuanOCR~\cite{hunyuanocr}        & arXiv 2025     & 77.80 & \underline{97.95} & \underline{3.13} & \underline{97.63} & \underline{98.06} & \underline{97.53} & \underline{95.07} & \underline{1.97} \\
MonkeyOCR-1.2B~\cite{monkeyocr}     & arXiv 2025     & -432.67 & 35.71 & 77.76 & 46.88 & 55.78 & 43.79 & 14.16 & 52.89\\
MonkeyOCR-3B~\cite{monkeyocr}       & arXiv 2025     & -184.49 & 53.07 & 58.15 & 62.00 & 66.40 & 59.59 & 28.88 &	46.35 \\
MinerU2.5~\cite{mineru25}           & arXiv 2025     &   48.68 & 58.65 & 47.35 & 66.93 & 62.78 & 75.14 & 44.60 & 39.44\\
PaddleOCR-VL-1.5~\cite{cui2026paddleocrvl15multitask09bvlm} & arXiv 2026  &   47.36 & 79.90 & 28.10 & 87.15 & 89.76 & 85.76 & 72.07 & 22.75 \\
MinerU2.5-Pro~\cite{mineru25pro}    & arXiv 2026     &   75.95 & 83.85 & 20.48 & 87.02 & 86.05 & 88.50 & 71.25 & 20.17\\
GLM-OCR~\cite{glmocr}               & arXiv 2026     & 69.47 & 74.56& 28.88 & 83.50 & 82.84 & 84.96 & 62.72 & 22.34 \\
PaddleOCR-VL-1.6~\cite{paddleocrvl16} &  arXiv 2026  &   72.25 & 89.08 & 15.59 & 92.27 & 93.36 & 92.19 & 85.04 & 14.29 \\
\textbf{TongGuOCR}                 & This work      & \textbf{97.93} & \textbf{98.33} & \textbf{2.05} & \textbf{98.03} & \textbf{98.07} & \textbf{98.02} & \textbf{96.23} & \textbf{1.51}  \\
\bottomrule
\end{tabular}
}
\vspace{2pt}
\begin{minipage}{0.98\textwidth}
\raggedright\scriptsize
\textit{Note.} Negative AR values occur when insertion errors are sufficiently severe that \(S+D+I>N\), indicating over-generation rather than a computational error; see Section~\ref{sec:evaluation_metrics}.
\end{minipage}
\end{table}

\begin{table}[!htbp]
\centering
\scriptsize
\setlength{\tabcolsep}{4pt}
\renewcommand{\arraystretch}{1.05}
\caption{Performance comparison on M\textsuperscript{5}HisDoc. AR, CR, NED, F1, P, R, BLEU, and RO-ED denote accuracy rate, correct rate, normalized edit distance, F1 score, precision, recall, BLEU, and reading-order edit distance, respectively. A dash denotes a result that was not reported. Higher values are better for AR, CR, F1, P, R, and BLEU, while lower values are better for NED and RO-ED. Bold and underlined values denote the best and second-best results among all compared methods, respectively. $^{\ast}$ indicates results reported under the text-line recognition setting and taken from the M\textsuperscript{5}HisDoc report~\cite{shi2023m5hisdoc}; in this setting, input line images are constructed from the dataset-provided text-line annotations.}
\label{tab:main_m5hisdoc}
\resizebox{\textwidth}{!}{%
\begin{tabular}{llcccccccc}
\toprule
\multirow{2}{*}{\textbf{Method}}
& \multirow{2}{*}{\textbf{Venue / Source}}
& \multicolumn{8}{c}{\textbf{M\textsuperscript{5}HisDoc}} \\
\cmidrule(lr){3-10}
& & \textbf{AR}$\uparrow$ & \textbf{CR}$\uparrow$ & \textbf{NED}$\downarrow$ & \textbf{F1}$\uparrow$ & \textbf{P}$\uparrow$ & \textbf{R}$\uparrow$ & \textbf{BLEU}$\uparrow$ & \textbf{RO-ED}$\downarrow$ \\
\midrule
\multicolumn{10}{l}{\textit{Traditional task-specific OCR models}} \\
CRNN~\cite{CRNN}$^{\ast}$ &  T-PAMI 2016 & 91.36 & 91.49 & - & - & - & - & - & -\\
Ma et al.~\cite{MTHv2}$^{\ast}$ & ICFHR 2020 & \underline{92.10} & \underline{92.29} & - & - & - & - & - & -\\
ZCTRN~\cite{ZCTRN}$^{\ast}$ &  ICDAR 2021 & 88.50 & 88.79 & - & - & - & - & - & - \\
ASTER~\cite{aster}$^{\ast}$ & T-PAMI 2018 & 87.70 & 87.98 & - & - & - & - & - & -\\
NRTR~\cite{nrtr}$^{\ast}$ &  ICDAR 2019& 83.02 & 85.43 & - & - & - & - & - & -\\
RobustScanner~\cite{robustscanner}$^{\ast}$ & ECCV 2020 & 90.78 & 90.99 & - & - & - & - & - & -\\
Peng et al.~\cite{peng2022recognition}$^{\ast}$ & TMM 2022 & 88.35 & 88.38 & - & - & - & - & - & -\\
HanDoc-OrderOCR~\cite{handoc_orderocr} & AAAI 2024& 54.45 & 55.71 & 45.45 & 65.68 & 73.17 & 63.18 & 40.42 & 32.59\\
Liu et al.~\cite{inner_engine}          & NSR 2023    &  89.47 & 91.07 & \underline{10.43} & \underline{92.02} & 91.23 & \underline{92.92} & \underline{83.74} & 7.71 \\
\midrule
\multicolumn{10}{l}{\textit{General-purpose MLLMs}} \\
Qwen3-VL-2B~\cite{qwen3vl}          & arXiv 2025     &  -100.41 & 49.37 & 63.55 & 59.56 & 77.77 & 54.51 & 29.42 & 	42.70 \\
Qwen3-VL-8B~\cite{qwen3vl}          & arXiv 2025     &  -5.07 & 64.64 & 45.11 & 72.48 & 81.19 & 69.48 & 45.86 & 32.64 \\
Qwen3.5-0.8B~\cite{qwen3.5}           & arXiv 2026 & -63.41 & 52.61 & 53.58 & 64.28 & 74.19 & 60.66 & 35.80 & 	39.47\\
Qwen3.5-9B~\cite{qwen3.5}           & arXiv 2026 & 53.01 & 67.42 & 35.80 & 75.73 & 79.86 & 73.54 & 51.86 & 30.02\\
InternVL3.5-2B~\cite{internvl3.5}   & arXiv 2025     & -130.28 & 37.82 & 68.75 & 58.50 & 82.82 & 50.94 & 28.32 & 58.80 \\
InternVL3.5-8B~\cite{internvl3.5}   & arXiv 2025     & -27.80 & 57.27 & 46.65 & 68.08 & 82.44 & 62.30 & 41.28 & 35.41 \\
Qwen3.7-Plus~\cite{qwen37plus}      & Official API & 64.55 & 70.77 & 32.86 & 76.73 & 80.02 & 75.27 & 55.92 & 24.86 \\
MiniMax M3~\cite{minimaxm3}      & Official API & 5.97 & 39.02 & 64.07 & 56.66 & 68.48 & 51.54 & 27.84 & 51.47 \\
Kimi K2.6~\cite{kimik26}      & Official API & 45.27 & 73.84 & 31.71 & 80.98 & 84.08 & 79.62 & 60.60  & 22.55 \\
GPT-5.5~\cite{gpt55}      & Official API & 45.01 & 51.34 & 51.88 & 63.88 & 70.70 & 61.05 & 39.33 & 46.45 \\
Gemini 3.1 Pro~\cite{gemini31pro}      & Official API & 61.05 & 67.76 & 35.99 & 76.41 & 79.68 & 74.58 & 53.80 & 29.33 \\

\midrule
\multicolumn{10}{l}{\textit{OCR-oriented MLLMs}} \\
DeepSeek-OCR 2~\cite{deepseekocr2}  & arXiv 2026     & -357.64 & 10.13 & 94.44 & 18.71 & 56.02 & 13.25 &  3.34 & 73.39 \\
DeepSeek-OCR~\cite{deepseekocr}     & arXiv 2025     & -208.84 & 37.65 & 70.68 & 53.45 & 73.03 & 46.85 & 23.38 & 53.87 \\
FireRed-OCR~\cite{fireredocr}       & arXiv 2026     & -244.73 & 61.03 & 46.42 & 68.48 & 79.03 & 65.26 & 43.39 & 26.67 \\
dots.ocr~\cite{dotsocr}             & arXiv 2025     &  -35.40 & 63.41 & 43.44 & 70.36 & 79.06 & 67.53 & 46.47 & 28.89 \\
Qianfan-OCR~\cite{dong2026qianfanocrunifiedendtoendmodel} & arXiv 2026 & 4.58 & 68.87 & 37.57 & 75.31 & 80.72 & 73.16 & 51.40 & 26.86 \\
HunyuanOCR~\cite{hunyuanocr}        & arXiv 2025     & 48.87 & 89.62 & 12.12 & 89.91 & \underline{91.27} & 89.18 & 79.91 & \underline{7.53} \\
MonkeyOCR-1.2B~\cite{monkeyocr}     & arXiv 2025     & -422.25 & 19.90 & 87.49 & 30.15 & 49.18 & 25.78 &  7.12 & 59.57 \\
MonkeyOCR-3B~\cite{monkeyocr}       & arXiv 2025     & -169.16 & 37.57 & 70.34 & 48.86 & 58.40 & 45.30 & 18.51 & 57.85 \\
MinerU2.5~\cite{mineru25}           & arXiv 2025     &   24.83 & 35.24 & 68.57 & 49.89 & 55.33 & 48.82 & 23.13 &  57.12 \\
PaddleOCR-VL-1.5~\cite{cui2026paddleocrvl15multitask09bvlm} & arXiv 2026  &   40.26 & 64.62 & 40.50 & 73.03 & 79.12 & 71.04 & 50.74 & 25.16\\
MinerU2.5-Pro~\cite{mineru25pro}    & arXiv 2026     &   38.48 & 60.20 & 45.75 & 69.96 & 70.52 & 70.79 & 45.39 & 35.39 \\
GLM-OCR~\cite{glmocr}               & arXiv 2026     & 46.73 & 55.73 & 48.76 & 66.40 & 66.88 & 68.38 & 41.86 & 34.06 \\
PaddleOCR-VL-1.6~\cite{paddleocrvl16} &  arXiv 2026  &   53.59 & 74.00 & 29.93 & 78.90 & 81.39 & 78.08 & 59.72 & 16.91 \\
\textbf{TongGuOCR}                 & This work      & \textbf{93.76} & \textbf{94.46} & \textbf{6.15} & \textbf{94.44} & \textbf{94.79} & \textbf{94.14} & \textbf{88.92} & \textbf{3.49} \\
\bottomrule
\end{tabular}
}
\vspace{2pt}
\begin{minipage}{0.98\textwidth}
\raggedright\scriptsize
\textit{Note.} Negative AR values occur when insertion errors are sufficiently severe that \(S+D+I>N\), indicating over-generation rather than a computational error; see Section~\ref{sec:evaluation_metrics}.
\end{minipage}
\end{table}

\noindent\textbf{Compared methods.}
The comparison includes representative methods from three categories: traditional task-specific OCR models, general-purpose MLLMs, and OCR-oriented MLLMs. Some traditional task-specific baselines use text-line crops constructed from dataset-provided annotations, whereas the MLLM-based methods operate on full-page images or use OCR-oriented processing pipelines.

\noindent\textbf{Results on MTHv2.}
As shown in Table~\ref{tab:main_mthv2}, on MTHv2, strong historical-document baselines already achieve high recognition accuracy, indicating that the benchmark is relatively saturated. TongGuOCR nevertheless reaches 97.93 AR, 98.33 CR, and 2.05 NED. Relative to the best competing score for each primary metric, it improves AR from 97.42 to 97.93 and CR from 97.95 to 98.33, while reducing NED from 3.13 to 2.05. It also achieves the best character-inventory metrics and BLEU score and reduces RO-ED to 1.51. These results suggest that TongGuOCR achieves the most accurate character recognition, the highest transcription fidelity, and the best preservation of reading order among all compared methods on MTHv2.

\noindent\textbf{Results on M\textsuperscript{5}HisDoc.}
As shown in Table~\ref{tab:main_m5hisdoc}, on M\textsuperscript{5}HisDoc, the advantage is more pronounced. This benchmark contains more diverse layouts, visual styles, backgrounds, and historical character forms. TongGuOCR achieves 93.76 AR, 94.46 CR, and 6.15 NED, compared with the best competing scores of 92.10, 92.29, and 10.43, respectively. It also obtains the best character-inventory metrics, BLEU score, and RO-ED. The reductions in NED from 10.43 to 6.15 and in RO-ED from 7.53 to 3.49 indicate better preservation of both content and reading order. These gains show that TongGuOCR is especially effective
on historical pages where layout complexity, difficult character
forms, and nontrivial reading paths jointly affect transcription
quality.

\noindent\textbf{Qualitative analysis.}
\begin{figure}[p]
    \centering
    \includegraphics[
        width=0.9\textwidth,
        height=0.86\textheight,
        trim=60 775 30 45,
        clip
    ]{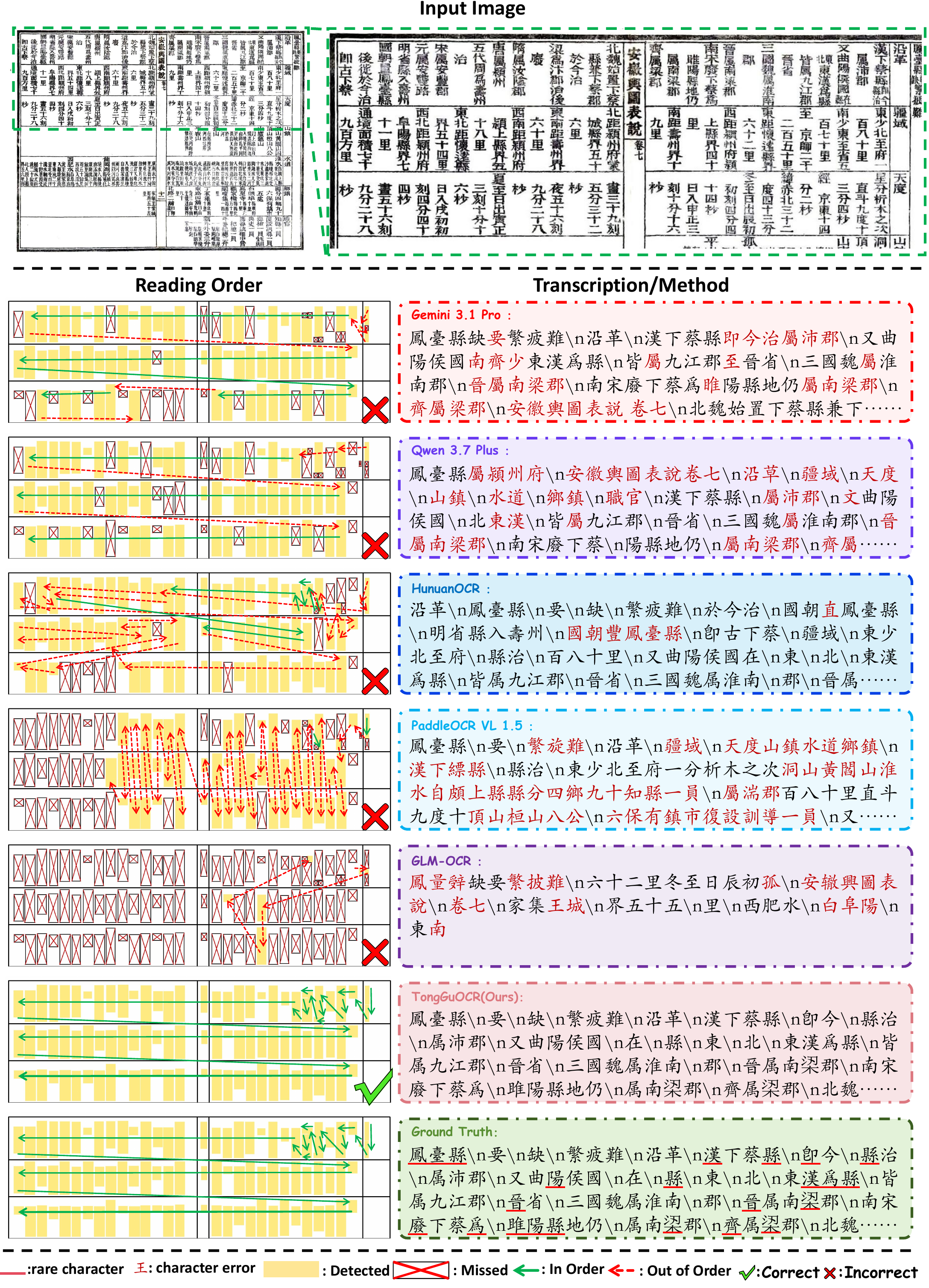}
    \caption{
Qualitative comparison on a challenging page from M\textsuperscript{5}HisDoc, which features a dense table-like layout and nontrivial reading-order transitions. The upper panel shows the input page and the enlarged target region. The lower-left panel visualizes the predicted reading paths, while the lower-right panel presents the corresponding transcriptions. Yellow boxes denote detected text lines, and crossed boxes indicate missed lines. Green and red arrows represent correct and incorrect reading-order transitions, respectively. Green checks and red crosses indicate whether the overall reading path is correct, while red underlining highlights rare or uncommon characters. Red characters, on the other hand, mark transcription errors. In this example, TongGuOCR recovers all target text lines, follows the natural reading sequence of the historical page, and accurately transcribes the corresponding content.
}
    \label{fig:main_comparison}
\end{figure}

Figure~\ref{fig:main_comparison} provides a qualitative comparison from the complementary perspectives of reading-path prediction and text transcription. In the reading-path visualization, the compared systems exhibit missed lines, interrupted paths, or out-of-order transitions. GPT-5.5 and Qwen3.7-Plus recover portions of the page content but fail to establish a reliable traversal sequence across the dense table-like layout, while HunyuanOCR, PaddleOCR-VL-1.5, and GLM-OCR also exhibit omissions or inconsistent transitions. Most of these systems are primarily optimized for modern document OCR and parsing. Although they retain some capability on historical materials, their predictions are less reliable on dense vertical layouts and nontrivial line transitions. In contrast, TongGuOCR recovers all target text lines and follows the natural reading sequence of the page.

The transcription panels show the consequences of the path errors described above. Missed text lines lead to omitted content, incorrect transitions place text segments in an unnatural sequence, and recognition errors alter individual characters. In contrast, TongGuOCR recovers every target line, preserves the natural reading sequence of the historical page, and accurately transcribes the corresponding content. This result provides a concrete illustration of the factors reflected in AR, NED, BLEU, and RO-ED: complete text recovery, accurate character recognition, and coherent line sequencing.

Overall, the results support the design of TongGuOCR, in which Layout-Aware Preprocessing provides structured and OCR-friendly recognition inputs, while Token-Augmented Recognition improves rare-character representation and provides spatial guidance during transcription through line-to-line transition modeling.

\subsection{Ablation Studies}

We organize the ablations around the two-module architecture and evaluate the staged SFT strategy separately. For Layout-Aware Preprocessing, we examine input-granularity alignment between SFT and inference and analyze the sensitivity to the segmentation penalty weights. For Token-Augmented Recognition, we ablate character-level vocabulary expansion and line-to-line transition modeling, followed by a detailed analysis of added and non-added character subsets. We then evaluate the contribution of each SFT stage.

All results are computed on the full official M\textsuperscript{5}HisDoc test split. We use this benchmark for ablation because its diverse layouts, visual styles, and historical character forms provide a more discriminative test than MTHv2~\cite{shi2023m5hisdoc}, on which the strongest baselines are already close to saturation (Table~\ref{tab:main_mthv2}). The module ablations exclude HisDoc1B because repeating full-parameter SFT on this large dataset for every variant would be prohibitively expensive. These variants are trained on the official training splits of MTHv2 and M\textsuperscript{5}HisDoc, with checkpoint selection on the official M\textsuperscript{5}HisDoc validation split. MTHv2 is used only for training in these ablations.

The staged SFT analysis follows the main training procedure and includes HisDoc1B in Stage 1. Its absolute scores should therefore not be compared with those in the module ablations, which use a different training-data configuration. Comparisons within each table remain controlled under the same protocol.

\subsubsection{Effect of Layout-Aware Preprocessing}
\begin{table}[!htbp]
\centering
\setlength{\tabcolsep}{7pt}
\renewcommand{\arraystretch}{1.35}
\caption{Effect of input-granularity alignment between SFT and inference on the M\textsuperscript{5}HisDoc test split. ``Page'' and ``Block'' denote full-page images and recognition-block crops, respectively.}
\label{tab:ablation_pipeline_m5hisdoc}
\fontsize{8pt}{10pt}\selectfont
\begin{adjustbox}{max width=0.9\textwidth}
\begin{tabular}{llcccccccc}
\toprule
\multirow{2}{*}{\textbf{SFT}}
& \multirow{2}{*}{\textbf{Infer.}}
& \multicolumn{8}{c}{\textbf{M\textsuperscript{5}HisDoc}} \\
\cmidrule(lr){3-10}
& & \textbf{AR}$\uparrow$ & \textbf{CR}$\uparrow$ & \textbf{NED}$\downarrow$ & \textbf{F1}$\uparrow$ & \textbf{P}$\uparrow$ & \textbf{R}$\uparrow$ & \textbf{BLEU}$\uparrow$  & \textbf{RO-ED}$\downarrow$\\
\midrule
Page  & Page  & \underline{83.50} & 86.66 & 14.00 & 84.86 & 85.98 & 84.03 & 73.37   & \underline{5.24} \\
Page  & Block & 75.43 & \underline{87.19} & \underline{13.93} & 85.55 & 86.15 & 85.08 & \underline{74.10} & 5.26 \\
Block & Page  & 68.47 & 74.11 & 29.83 & \underline{86.34} & \underline{88.10} & \underline{85.44} & 72.61 & 26.60\\
Block & Block & \textbf{89.69} & \textbf{90.88} & \textbf{9.79} & \textbf{90.17} & \textbf{90.32} & \textbf{90.07} & \textbf{82.30} & \textbf{4.93}\\
\bottomrule
\end{tabular}
\end{adjustbox}
\end{table}

\begin{table}[!htbp]
\centering
\setlength{\tabcolsep}{5pt}
\renewcommand{\arraystretch}{1.25}
\caption{Sensitivity analysis of penalty weights in Recognition Block Construction on the M\textsuperscript{5}HisDoc test split. 
Each setting is represented as \((\lambda_{\mathrm{blk}}, \lambda_{\mathrm{adj}}, \lambda_{\mathrm{page}})\), where the three coefficients weight the block-level, adjacent-block, and page-level penalties, respectively. 
\(\dagger\) denotes the default setting used in our main ablation experiments.}
\label{tab:ablation_layout_weight_m5hisdoc}
\fontsize{8pt}{10pt}\selectfont
\begin{adjustbox}{max width=0.9\textwidth}
\begin{tabular}{lcccccccc}
\toprule
\multirow{2}{*}{\textbf{Setting}}
& \multicolumn{8}{c}{\textbf{M\textsuperscript{5}HisDoc}} \\
\cmidrule(lr){2-9}
& \textbf{AR}$\uparrow$ 
& \textbf{CR}$\uparrow$ 
& \textbf{NED}$\downarrow$ 
& \textbf{F1}$\uparrow$ 
& \textbf{P}$\uparrow$ 
& \textbf{R}$\uparrow$ 
& \textbf{BLEU}$\uparrow$ 
& \textbf{RO-ED}$\downarrow$ \\
\midrule
\((1,1,1)\) & 85.44 & 90.06 & 10.75 & 89.18 & 89.31 & 89.11 & 80.46 & 5.29 \\
\((2,1,1)\) & 85.94 & 89.92 & 11.06 & 89.33 & 89.85 & 89.02 & 80.47 & 5.38 \\
\((1,2,1)\) & 86.75 & \underline{90.17} & \underline{10.66} & \underline{89.55} & \underline{89.92} & \underline{89.29} & \underline{80.87} & 5.10 \\
\((4,2,1)\) & \underline{87.96} & 89.89 & 10.82 & 89.16 & 89.51 & 88.92 & 80.50 & \underline{5.07} \\
\((2,4,1)\) & 86.93 & 89.74 & 10.95 & 89.09 & 89.62 & 88.78 & 80.35 & 5.48 \\
\((4,4,1)\) & 84.57 & 88.91 & 11.87 & 87.95 & 88.46 & 87.64 & 78.54 & 5.34 \\
\midrule
\((2,2,1)^{\dagger}\) & \textbf{89.69} & \textbf{90.88} & \textbf{9.79} & \textbf{90.17} & \textbf{90.32} & \textbf{90.07} & \textbf{82.30} & \textbf{4.93} \\
\bottomrule
\end{tabular}
\end{adjustbox}
\end{table}
\noindent\textbf{Effectiveness of Recognition Blocks.}
This experiment evaluates whether recognition blocks improve OCR performance over full-page inputs and whether they should be used consistently during both SFT and inference. Page-level SFT uses full-page samples, whereas block-level SFT uses recognition-block crops. Inference is likewise performed with either full pages or recognition blocks. Page/Page and Block/Block therefore use matched input granularities, while Page/Block and Block/Page introduce a mismatch. All four configurations retain character-level vocabulary expansion and line-to-line transition modeling.

Table~\ref{tab:ablation_pipeline_m5hisdoc} shows that Block/Block performs best on every metric, achieving 89.69 AR, 90.88 CR, 82.30 BLEU, and 4.93 RO-ED. Compared with Page/Page, it improves AR by 6.19 points and BLEU by 8.93 points, while reducing NED from 14.00 to 9.79. These results suggest that recognition blocks are most effective when used consistently during both SFT and inference. Because this comparison changes the input granularity at both stages, it supports the complete recognition-block interface rather than an isolated test-time preprocessing effect.

The remaining configurations perform less consistently for different reasons. Page/Block introduces recognition blocks only during inference, even though the model is trained with full-page images. Although CR increases from 86.66 to 87.19 and BLEU from 73.37 to 74.10, AR decreases from 83.50 to 75.43, showing that this test-time change does not provide a consistent benefit. Block/Page creates the opposite mismatch and performs worst, with 68.47 AR and 26.60 RO-ED. Its F1 remains relatively high at 86.34, but this set-based metric does not capture the severe errors in complete transcription and reading order. Page/Page avoids a granularity mismatch but still underperforms Block/Block, a pattern consistent with the loss of character detail and cross-region interference associated with full-page inputs. Overall, the results support the consistent use of recognition blocks during both SFT and inference.

\noindent\textbf{Penalty-weight sensitivity.}
Based on the aligned Block/Block configuration, Table~\ref{tab:ablation_layout_weight_m5hisdoc} evaluates the relative weights of the three penalty groups in Recognition Block Construction. Since multiplying all three weights by the same positive constant does not change the optimal segmentation, we fix \(\lambda_{\mathrm{page}}=1\) as the reference and vary \(\lambda_{\mathrm{blk}}\) and \(\lambda_{\mathrm{adj}}\). The experiments therefore compare different relative weight ratios rather than their absolute magnitudes.

The default ratio (2,2,1) performs best on every metric, achieving 89.69 AR, 9.79 NED, and 4.93 RO-ED. The ratios (4,2,1) and (1,2,1) remain competitive on different metrics, but neither matches the default setting. Increasing both local penalty weights to (4,4,1) reduces AR to 84.57 and raises NED to 11.87. This decline suggests that excessive emphasis on block-level and adjacent-block constraints can overconstrain the segmentation. The default ratio therefore provides the best balance among block quality, adjacent-block compatibility, and page-level segmentation complexity.

\subsubsection{Effect of Token-Augmented Recognition}
\begin{table}[!htbp]
\centering
\setlength{\tabcolsep}{8pt}
\renewcommand{\arraystretch}{1.55}
\caption{Ablation of character-level vocabulary expansion and line-to-line transition modeling on the M\textsuperscript{5}HisDoc test split. ``Vocab.'' denotes character-level vocabulary expansion, and ``Trans.'' denotes line-to-line transition modeling.}
\label{tab:ablation_token_m5hisdoc}
\fontsize{8pt}{10pt}\selectfont
\begin{adjustbox}{max width=0.9\textwidth}
\begin{tabular}{cccccccccc}
\toprule
\multirow{2}{*}{\textbf{Vocab.}}
& \multirow{2}{*}{\textbf{Trans.}}
& \multicolumn{8}{c}{\textbf{M\textsuperscript{5}HisDoc}} \\
\cmidrule(lr){3-10}
& & \textbf{AR}$\uparrow$ & \textbf{CR}$\uparrow$ & \textbf{NED}$\downarrow$ & \textbf{F1}$\uparrow$ & \textbf{P}$\uparrow$ & \textbf{R}$\uparrow$ & \textbf{BLEU}$\uparrow$  & \textbf{RO-ED}$\downarrow$ \\
\midrule
-- & -- & 80.89 & 89.67 & 11.23 & 88.92 & 89.12 & 89.05 & 80.44 & 5.56 \\
\checkmark & -- & 86.12 & \underline{90.13} & \underline{10.65} & \underline{89.47} & \underline{90.01} & 89.23 & \underline{81.10} & \underline{5.23}\\
-- & \checkmark & \underline{87.89} & 90.12 & \underline{10.65} & 89.18 & 89.14 & \underline{89.36} & 80.94 & 5.24 \\
\checkmark & \checkmark & \textbf{89.69} & \textbf{90.88} & \textbf{9.79} & \textbf{90.17} & \textbf{90.32} & \textbf{90.07} & \textbf{82.30} & \textbf{4.93} \\
\bottomrule
\end{tabular}
\end{adjustbox}
\end{table}

\begin{table}[!htbp]
\centering
\setlength{\tabcolsep}{4.5pt}
\renewcommand{\arraystretch}{1.25}
\caption{Occurrence-level character alignment analysis of token-augmented recognition on the M\textsuperscript{5}HisDoc test split. Characters are divided into Added chars and Non-added chars according to whether they belong to the newly introduced single-character tokens. ``Vocab.'' denotes character-level vocabulary expansion, and ``Trans.'' denotes line-to-line transition modeling. FP and FN denote over-generated and missed character occurrences after full-text character-level alignment, respectively.}
\label{tab:rare_char_alignment}
\fontsize{8pt}{10pt}\selectfont
\begin{adjustbox}{max width=0.9\textwidth}
\begin{tabular}{cc ccc rr ccc}
\toprule
\multirow{2}{*}{\textbf{Vocab.}}
& \multirow{2}{*}{\textbf{Trans.}}
& \multicolumn{5}{c}{\textbf{Added chars}} 
& \multicolumn{3}{c}{\textbf{Non-added chars}} \\
\cmidrule(lr){3-7} \cmidrule(lr){8-10}
& & \textbf{P}$\uparrow$ & \textbf{R}$\uparrow$ & \textbf{F1}$\uparrow$
& \textbf{FP}$\downarrow$ & \textbf{FN}$\downarrow$
& \textbf{P}$\uparrow$ & \textbf{R}$\uparrow$ & \textbf{F1}$\uparrow$ \\
\midrule
-- & -- & 75.17 & 78.73 & 76.91 & 33270 & 27219 & 79.26 & 91.21 & 84.82 \\
\checkmark & -- & \underline{84.39} & 78.43 & \underline{81.30} & \underline{18567} & 27602 & 84.55 & \underline{92.17} & 88.19 \\
-- & \checkmark & 81.66 & \underline{79.21} & 80.42 & 22758 & \underline{26605} & \underline{88.56} & 92.00 & \underline{90.25} \\
\checkmark & \checkmark & \textbf{85.43} & \textbf{80.42} & \textbf{82.85} & \textbf{17547} & \textbf{25057} & \textbf{91.42} & \textbf{92.80} & \textbf{92.11} \\
\bottomrule
\end{tabular}
\end{adjustbox}
\end{table}
This ablation evaluates the two token-level designs introduced in Sections~\ref{sec:vocab_expansion} and~\ref{sec:line_transition}: character-level vocabulary expansion and line-to-line transition modeling. All variants use recognition-block inputs during both SFT and inference, following the aligned setting established in the previous ablation, and only these two token-level designs are changed.

\noindent\textbf{Effects of Vocabulary Expansion and Transition Modeling.}
Table~\ref{tab:ablation_token_m5hisdoc} shows that both designs improve the baseline and provide complementary benefits. Vocabulary expansion raises AR from 80.89 to 86.12 and reduces NED from 11.23 to 10.65, indicating more stable character generation after reducing token fragmentation. Transition modeling raises AR to 87.89, which suggests that explicit spatial cues help the decoder move from a completed line to the next relevant region. The two single-component variants achieve comparable RO-ED scores. Their combination performs best on every metric, reaching 89.69 AR, 9.79 NED, 82.30 BLEU, and 4.93 RO-ED.

The transition-only variant obtains an RO-ED of 5.24, which is slightly higher than the 5.23 achieved by vocabulary expansion alone. This small irregularity may arise because RO-ED is computed after predicted lines are matched to reference lines according to textual similarity. Recognition errors can therefore affect line matching and the resulting order score. The 0.01 difference does not contradict the overall benefit of transition modeling, which improves AR from 80.89 to 87.89.

\noindent\textbf{Performance on Added and Non-Added Characters.}
Table~\ref{tab:rare_char_alignment} shows how the two designs affect different character subsets. Added chars are occurrences represented by the new single-character tokens, while Non-added chars are the remaining normalized occurrences. The baseline obtains F1 scores of 76.91 and 84.82 on these subsets, respectively, confirming that the added characters are more difficult.

Vocabulary expansion primarily benefits the added-character subset. Without transition modeling, it raises precision from 75.17 to 84.39 and F1 from 76.91 to 81.30. Recall decreases slightly from 78.73 to 78.43, but false positives fall from 33,270 to 18,567. The expanded vocabulary therefore makes predictions more precise with little loss in coverage.

Transition modeling provides a different benefit. Without vocabulary expansion, it raises F1 from 84.82 to 90.25 for non-added characters and from 76.91 to 80.42 for added characters. This result suggests that better sequence organization reduces alignment errors caused by complex layouts. Combining both designs produces the best F1 on each subset, with scores of 82.85 for added characters and 92.11 for non-added characters. Vocabulary expansion improves difficult character representations, while transition modeling stabilizes the output sequence.

\subsubsection{Effect of Staged Supervised Fine-Tuning}
\begin{table}[!htbp]
\centering
\setlength{\tabcolsep}{4.5pt}
\renewcommand{\arraystretch}{1.55}
\caption{Effect of staged SFT on the M\textsuperscript{5}HisDoc test split. Stage 1 denotes coarse domain adaptation, Stage 2 denotes high-quality page-level refinement, and Stage 3 denotes block-level inference alignment. The reported rows correspond to the final Stage 1 checkpoint at epoch 3 and the validation-selected Stage 2 and Stage 3 checkpoints at epochs 5 and 3, respectively.}
\label{tab:ablation_stage_m5hisdoc}
\fontsize{8pt}{10pt}\selectfont
\begin{adjustbox}{max width=0.9\textwidth}
\begin{tabular}{ccccccccccc}
\toprule
\multirow{2}{*}{\textbf{Stage 1}}
& \multirow{2}{*}{\textbf{Stage 2}}
& \multirow{2}{*}{\textbf{Stage 3}}
& \multicolumn{8}{c}{\textbf{M\textsuperscript{5}HisDoc}} \\
\cmidrule(lr){4-11}
& & & \textbf{AR}$\uparrow$ & \textbf{CR}$\uparrow$ & \textbf{NED}$\downarrow$ & \textbf{F1}$\uparrow$ & \textbf{P}$\uparrow$ & \textbf{R}$\uparrow$ & \textbf{BLEU}$\uparrow$  & \textbf{RO-ED}$\downarrow$\\
\midrule
\checkmark & -- & -- & \underline{90.29} & 91.55 & 9.57 & 93.08 & 92.95 & 93.28 & 84.32 & 6.42	\\
\checkmark & \checkmark & -- & 90.03 & \underline{93.56} & \underline{7.47} & \underline{93.75} & \underline{94.18} & \underline{93.42} & \underline{86.93} & \underline{4.07} \\
\checkmark & \checkmark & \checkmark & \textbf{93.76} & \textbf{94.46} & \textbf{6.15} & \textbf{94.44} & \textbf{94.79} & \textbf{94.14} & \textbf{88.92} & \textbf{3.49}\\
\bottomrule
\end{tabular}
\end{adjustbox}
\end{table}
This ablation examines whether the three SFT stages fulfill their intended roles in adapting domain knowledge, improving supervision quality, and aligning input granularity. The three rows correspond to coarse domain adaptation, high-quality page-level refinement, and block-level inference alignment. All models use recognition-block inputs during evaluation.

Table~\ref{tab:ablation_stage_m5hisdoc} shows that Stage 1 establishes a strong historical-document OCR foundation through large-scale domain adaptation. Stage 2 then uses human-annotated page-level data for high-quality refinement, improving CR from 91.55 to 93.56, reducing NED from 9.57 to 7.47, increasing BLEU from 84.32 to 86.93, and reducing RO-ED from 6.42 to 4.07. These improvements are consistent with its intended role in refining character recognition and overall sequence fidelity. AR decreases only slightly from 90.29 to 90.03; since AR additionally penalizes insertions while all other reported metrics improve, this minor exception reflects increased insertion errors rather than an overall performance degradation. These additional insertions may partly arise from the remaining granularity mismatch between page-level training and recognition-block inference.

Stage 3 aligns training with inference by using recognition blocks generated through the same preprocessing procedure used at test time. It improves every metric, raising AR from 90.03 to 93.76 and CR from 93.56 to 94.46, while reducing NED from 7.47 to 6.15 and RO-ED from 4.07 to 3.49. AR improves more than CR, and the gap between CR and AR decreases from 3.53 to 0.70. These changes indicate fewer insertion errors and support input-granularity alignment. Although Stages 2 and 3 are scheduled for 10 and 15 epochs, their validation-selected checkpoints occur at epochs 5 and 3, respectively. Continuing either stage produces no better checkpoint. The gains therefore reflect the intended changes in supervision quality and input granularity rather than longer training alone.

\section{Discussion}

The results show that reliable OCR for Chinese historical documents requires coordinated treatment of layout, character representation, and reading order. Recognition blocks preserve character visibility while retaining local context. Character-level vocabulary expansion creates a more direct correspondence between visual glyphs and output tokens for rare characters. Transition tokens provide spatial supervision at line boundaries and help the decoder maintain a coherent reading path. Together, these designs achieve 93.76 AR and 3.49 RO-ED on M\textsuperscript{5}HisDoc, reflecting strong character recognition and reading-order preservation.

The strong performance of TongGuOCR across the evaluated benchmarks further demonstrates the effectiveness of this design. Compared with existing approaches, TongGuOCR achieves more accurate and stable transcription for Chinese historical documents, especially in scenarios where layout complexity and rare characters substantially affect recognition quality. These results suggest that explicitly modeling layout structure and historical character distributions is important for improving OCR performance on historical materials. Beyond benchmark accuracy, TongGuOCR also has practical value for digital humanities and cultural heritage preservation, where reliable transcription can support full-text retrieval, textual collation, named-entity extraction, knowledge organization, and large-scale access to image-based historical collections.

TongGuOCR still has two limitations. First, the expanded vocabulary is derived from the training corpus and does not cover all historical characters, including those absent from the training data or not yet deciphered. Second, the evaluated datasets do not represent the full diversity of Chinese historical materials. Future work will investigate more flexible representations for unseen characters and evaluate TongGuOCR on more diverse historical materials.

\section{Conclusion}

In this paper, we present TongGuOCR, a new OCR MLLM designed for the distinctive challenges of Chinese historical documents. Faithful OCR of Chinese historical documents requires coordinated modeling of recognition granularity, character representation, and cross-line structure. Our results show that recognition blocks are most effective when training and inference use the same granularity, and that direct character-level representations and explicit line-boundary supervision provide complementary benefits for rare-character fidelity and multi-line sequence organization. These findings establish a broader design principle: layout-guided inputs and structured transcription targets should be designed jointly rather than as isolated components. TongGuOCR validates this principle across representative benchmarks and provides a stronger foundation for retrieval, textual collation, and computational study of Chinese historical collections.

\clearpage
\bibliographystyle{unsrtnat}
\bibliography{references}

\clearpage
\appendix
\setcounter{section}{0}
\setcounter{subsection}{0}
\setcounter{equation}{0}
\setcounter{table}{0}
\renewcommand{\thesection}{S\arabic{section}}
\renewcommand{\thesubsection}{\thesection.\arabic{subsection}}
\renewcommand{\theequation}{S\arabic{equation}}
\renewcommand{\thetable}{S\arabic{table}}
\section*{Supplementary Material}
\section{Additional Details of Layout-Aware Preprocessing}

This Supplementary Note provides additional implementation details for the Layout-Aware Preprocessing module described in the main manuscript. The main manuscript describes the overall design, the penalty-based segmentation objective, and the role of Layout-Aware Preprocessing in TongGuOCR. This note gives further details on candidate block generation, penalty construction, dynamic programming, and Block Crop Refinement.

\subsection{Overview}

Layout-Aware Preprocessing converts an ordered text-line sequence into an ordered sequence of OCR-friendly crop images. It contains two main stages. The first stage, Recognition Block Construction, groups ordered text lines into recognition blocks by solving a layout-guided sequence segmentation problem. The second stage, Block Crop Refinement, converts each selected block into an OCR-friendly crop image by crop expansion, non-target text masking, local background filling, and square-canvas padding.

Given an ordered text-line sequence
\[
L=[l_1,l_2,\ldots,l_n],
\]
each text line \(l_i\) is associated with a bounding box \(b_i\). The goal is to partition \(L\) into a sequence of recognition blocks
\[
S=[B_1,B_2,\ldots,B_K],
\]
where each \(B_t\) is a consecutive subsequence of \(L\). This interval-based representation provides a structured search space for block construction and supports efficient global sequence segmentation.

\subsection{Candidate Block Generation}

Candidate recognition blocks are generated as consecutive intervals:
\begin{equation}
B_{i,j}=L[i:j], \quad 1\leq i<j\leq n+1 .
\end{equation}
Here, \(B_{i,j}\) contains text lines \(l_i,l_{i+1},\ldots,l_{j-1}\). The bounding box of a candidate block is defined as the minimum bounding rectangle (MBR) covering all text-line boxes in the interval:
\begin{equation}
\operatorname{BBox}(B_{i,j})
=
\operatorname{MBR}\!\left(\left\{b_k \mid i\leq k<j\right\}\right),
\quad 1\leq i<j\leq n+1 .
\end{equation}

Candidate enumeration is restricted by maximum interval length and extreme geometry pruning. The pruning step removes candidates that are clearly unsuitable for OCR, such as extremely large regions, very thin strip-like regions, and candidates with severe geometric inconsistency. This reduces the search space while keeping candidates that may form useful OCR inputs.

\subsection{Penalty-Based Segmentation Objective}

Recognition Block Construction is formulated as a minimum-penalty sequence segmentation problem. For a valid segmentation \(S=[B_1,\ldots,B_K]\), the total penalty is
\begin{equation}
\mathcal{P}(S)
=
\lambda_{\mathrm{blk}}\mathcal{P}_{\mathrm{blk}}(S)
+
\lambda_{\mathrm{adj}}\mathcal{P}_{\mathrm{adj}}(S)
+
\lambda_{\mathrm{page}}\mathcal{P}_{\mathrm{page}}(K).
\end{equation}
The three penalty groups correspond to candidate-block quality, adjacent-block compatibility, and page-level block-number regularization. Larger penalty values indicate less desirable candidates or segmentations.

The block-level penalty group aggregates single-block penalties over all selected blocks:
\begin{equation}
\mathcal{P}_{\mathrm{blk}}(S)
=
\sum_{t=1}^{K}p_{\mathrm{blk}}(B_t).
\end{equation}
The adjacent-block penalty group aggregates compatibility penalties over neighboring block pairs:
\begin{equation}
\mathcal{P}_{\mathrm{adj}}(S)
=
\sum_{t=1}^{K-1}p_{\mathrm{adj}}(B_t,B_{t+1}).
\end{equation}
The page-level penalty depends on the number of selected blocks:
\begin{equation}
\mathcal{P}_{\mathrm{page}}(K)=p_{\mathrm{page}}(K).
\end{equation}

\subsection{Block-Level Penalty}

The single-block penalty \(p_{\mathrm{blk}}(B)\) evaluates whether a candidate interval \(B\) forms an OCR-friendly recognition block. It combines geometric and layout cues related to character scale, local context, text compactness, crop shape, and line consistency:
\begin{equation}
\begin{aligned}
p_{\mathrm{blk}}(B)
={}&
\omega_n P_n(B)
+
\omega_a P_a(B)
+
\omega_f P_f(B) \\
&+
\omega_r P_r(B)
+
\omega_s P_s(B)
+
\omega_g P_g(B) \\
&+
\omega_o P_o(B)
+
\omega_{\Delta}P_{\Delta}(B).
\end{aligned}
\end{equation}
Here, \(P_n,\ldots,P_{\Delta}\) are normalized penalty terms, and all \(\omega\) terms are non-negative weights. These terms denote semantic groups of geometric cues; in the implementation, a group may contain both soft and hard penalty components. The individual penalty terms are summarized in Supplementary Table~\ref{tab:supp_block_penalty}.

\begin{table}[!htbp]
\centering
\caption{\textbf{Block-level penalty terms.} Each term measures one property of a candidate recognition block.}
\label{tab:supp_block_penalty}
\scriptsize
\setlength{\tabcolsep}{5pt}
\renewcommand{\arraystretch}{1.18}
\begin{tabularx}{0.95\textwidth}{p{0.14\textwidth}p{0.23\textwidth}X}
\toprule
\textbf{Term} & \textbf{Cue} & \textbf{Penalized case} \\
\midrule
\(P_n(B)\) & Number of text lines & The candidate contains too few lines, which produces fragmented inputs, or too many lines, which makes the crop text-rich and difficult to recognize. \\
\(P_a(B)\) & Relative block area & The candidate covers an overly large region of the page and may reduce character scale after resizing. \\
\(P_f(B)\) & Text fill ratio & The candidate contains large blank regions or combines spatially inconsistent lines. \\
\(P_r(B)\) & Aspect ratio & The candidate has an overly elongated shape. \\
\(P_s(B)\) & Strip-like geometry & The candidate forms a narrow strip-like crop that is unstable after square padding. \\
\(P_g(B)\) & Horizontal gap & Neighboring lines inside the candidate have large horizontal gaps. \\
\(P_o(B)\) & Vertical overlap & Neighboring lines inside the candidate have low vertical overlap, often indicating a transition to another local region. \\
\(P_{\Delta}(B)\) & Block growth stability & Adding a line causes abnormal expansion of the candidate region or a large change in its stable vertical band. \\
\bottomrule
\end{tabularx}
\end{table}

Let \(m(B)\) denote the number of text lines in \(B\). Let
\[
A(B)=\frac{\operatorname{area}(\operatorname{BBox}(B))}{\operatorname{area}(I)}
\]
denote the relative area of the candidate block in the page image \(I\). The line-count term \(P_n(B)\) penalizes candidates whose line count is outside the preferred range, while the area term \(P_a(B)\) penalizes overly large candidates.

The text fill ratio used by \(P_f(B)\) is defined as
\begin{equation}
\operatorname{fill}(B)=
\frac{\sum_{l_i \in B} \operatorname{area}(b_i)}
{\operatorname{area}(\operatorname{BBox}(B))}.
\end{equation}
A low fill ratio indicates that the candidate contains too much blank area or combines lines from spatially inconsistent regions. The corresponding penalty \(P_f(B)\) increases as the fill ratio decreases.

The aspect-ratio term \(P_r(B)\) penalizes candidate boxes with highly imbalanced width and height. If the width and height of \(\operatorname{BBox}(B)\) are \(w_B\) and \(h_B\), the aspect-ratio imbalance is measured as
\begin{equation}
r(B)=\max\left(\frac{w_B}{h_B},\frac{h_B}{w_B}\right).
\end{equation}
The strip-like term \(P_s(B)\) further penalizes narrow strip-like crops, which are not suitable as OCR inputs after square padding.

The line-consistency terms \(P_g(B)\) and \(P_o(B)\) are computed from neighboring text-line boxes inside the candidate interval. Horizontal gaps are normalized by the page width. Vertical overlap is measured between neighboring line boxes and normalized by the smaller line height. For vertical historical document pages, lines in the same local region usually share a stable vertical band. Therefore, a sharp drop in vertical overlap suggests that the candidate is crossing into another column, table-like region, or marginal area.

The block-growth term \(P_{\Delta}(B)\) is computed during interval expansion. It penalizes candidates whose bounding box changes abruptly when a new line is appended, such as a large increase in area or a large shift of the stable vertical band. This helps prevent unrelated lines from being merged into the same recognition block.

\subsection{Adjacent-Block Penalty}

The single adjacent-block penalty \(p_{\mathrm{adj}}(B,B')\) evaluates the geometric compatibility between two neighboring recognition blocks \(B\) and \(B'\). Let
\[
R=\operatorname{BBox}(B), \qquad R'=\operatorname{BBox}(B').
\]
The penalty is computed from geometric relations between \(R\) and \(R'\):
\begin{equation}
\begin{aligned}
p_{\mathrm{adj}}(B,B')
={}&
\gamma_{\mathrm{ov}}P_{\mathrm{ov}}(B,B')
+
\gamma_{\mathrm{con}}P_{\mathrm{con}}(B,B') \\
&+
\gamma_{\mathrm{tiny}}P_{\mathrm{tiny}}(B,B') \\
&+
\gamma_{\mathrm{split}}P_{\mathrm{split}}(B,B').
\end{aligned}
\end{equation}
Here, \(P_{\mathrm{ov}}, P_{\mathrm{con}}, P_{\mathrm{tiny}}\), and \(P_{\mathrm{split}}\) are normalized geometric penalty terms, and all \(\gamma\) terms are non-negative weights. The adjacent-block penalty terms are summarized in Supplementary Table~\ref{tab:supp_adj_penalty}.

\begin{table}[!htbp]
\centering
\caption{\textbf{Adjacent-block penalty terms.} Each term measures one geometric relation between neighboring block boxes.}
\label{tab:supp_adj_penalty}
\scriptsize
\setlength{\tabcolsep}{5pt}
\renewcommand{\arraystretch}{1.18}
\begin{tabularx}{0.95\textwidth}{p{0.18\textwidth}p{0.23\textwidth}X}
\toprule
\textbf{Term} & \textbf{Cue} & \textbf{Penalized case} \\
\midrule
\(P_{\mathrm{ov}}\) & Box overlap & Neighboring block boxes overlap excessively. \\
\(P_{\mathrm{con}}\) & Containment relation & One neighboring block box is largely contained by the other. \\
\(P_{\mathrm{tiny}}\) & Relative size imbalance & A tiny block is embedded in or attached to a much larger neighboring region. \\
\(P_{\mathrm{split}}\) & Split-boundary stability & The split creates unstable neighboring crops or separates highly related line groups. \\
\bottomrule
\end{tabularx}
\end{table}

The overlap term \(P_{\mathrm{ov}}\) is based on the intersection-over-union or intersection ratio between neighboring block boxes. The containment term \(P_{\mathrm{con}}\) penalizes cases where one block is largely inside another block. The relative-size term \(P_{\mathrm{tiny}}\) penalizes tiny blocks attached to or embedded in much larger neighboring regions. The split-boundary term \(P_{\mathrm{split}}\) discourages unstable boundaries between closely related line groups.

These terms reduce cross-block interference and discourage unstable partitions. They are especially useful in dense vertical pages, table-like layouts, and pages with marginal annotations, where spatial proximity alone may lead to unsuitable grouping.

\subsection{Page-Level Penalty}

The page-level penalty \(p_{\mathrm{page}}(K)\), where \(K\) is the number of selected recognition blocks, regularizes the number of recognition blocks on a page. It discourages both excessive fragmentation and the degenerate solution of using the entire page as a single block. We first define the normalized quadratic deviation from the preferred range \([K_{\min},K_{\max}]\) as
\begin{equation}
\begin{aligned}
&q_{\mathrm{range}}(K)=\\[-2pt]
&\quad
\begin{cases}
\left(\dfrac{K_{\min}-K}{K_{\min}}\right)^2,
& K<K_{\min},\\[6pt]
0,
& K_{\min}\leq K\leq K_{\max},\\[6pt]
\left(\dfrac{K-K_{\max}}{K_{\max}}\right)^2,
& K>K_{\max}.
\end{cases}
\end{aligned}
\end{equation}
An additional quadratic term is applied only when the block count exceeds a soft upper cap \(K_{\mathrm{soft}}>K_{\max}\):
\begin{equation}
q_{\mathrm{cap}}(K)
=
\left(
\frac{[K-K_{\mathrm{soft}}]_+}{K_{\mathrm{soft}}}
\right)^2.
\end{equation}
The page-level penalty is then
\begin{equation}
p_{\mathrm{page}}(K)
=
\rho_{\mathrm{range}}q_{\mathrm{range}}(K)
+
\rho_{\mathrm{cap}}q_{\mathrm{cap}}(K)
+
\rho_{\mathrm{single}}\mathbb{I}(K=1).
\end{equation}
Here, \([x]_+=\max(x,0)\), \(\mathbb{I}(\cdot)\) is the indicator function, and all \(\rho\) terms are non-negative weights. The preferred-range term penalizes block counts below \(K_{\min}\) or above \(K_{\max}\), the soft-cap term strengthens the penalty for severe fragmentation, and the single-block term prevents the optimization from choosing one whole-page block, which would weaken the purpose of Layout-Aware Preprocessing.

\subsection{Dynamic Programming Solution}

After candidate intervals are enumerated, dynamic programming is used to find the minimum-penalty block sequence. Let \(D(B_{i,j},k)\) denote the minimum accumulated penalty for a partial segmentation that ends with candidate block \(B_{i,j}\) and contains \(k\) blocks. The recurrence is
\begin{equation}
\begin{aligned}
D(B_{i,j},k)
={}&
\lambda_{\mathrm{blk}}p_{\mathrm{blk}}(B_{i,j}) \\
&+
\min_{h<i}
\left[
D(B_{h,i},k-1) \right. \\
&\qquad \left. +\lambda_{\mathrm{adj}}p_{\mathrm{adj}}(B_{h,i},B_{i,j})
\right].
\end{aligned}
\end{equation}
Here, the minimization is over valid predecessor intervals \(B_{h,i}\) that end immediately before \(B_{i,j}\). The terminal state is selected by adding the page-level penalty:
\begin{equation}
(k^*,B_{\mathrm{end}}^*)
=
\arg\min_{k,B_{i,n+1}}
\left[
D(B_{i,n+1},k)
+
\lambda_{\mathrm{page}}p_{\mathrm{page}}(k)
\right].
\end{equation}
The minimum-penalty recognition block sequence within the retained candidate space is recovered by backtracking from the selected terminal block \(B_{\mathrm{end}}^*\). We then apply a lightweight local merge step to suppress isolated tiny blocks. A tiny block is considered for merging with either adjacent block, and the merged interval is retained only when its geometry remains suitable for recognition according to fill-ratio, aspect-ratio, and strip-like constraints. The local decision is evaluated using the same block-level and adjacent-block costs as the sequence-segmentation objective. This postprocessing step reduces unstable fragmentation while preserving the order and consecutive coverage of the text-line sequence.

\subsection{Block Crop Refinement}

After Recognition Block Construction, each selected block is converted into an OCR-friendly crop image. For a block \(B_t\), we first compute its union bounding box \(\operatorname{BBox}(B_t)\) and expand it by a fixed margin \(p\):
\begin{equation}
\Omega_t = \operatorname{Expand}(\operatorname{BBox}(B_t), p).
\end{equation}
The expanded crop is clipped to the page boundary.

Because the expanded crop may include lines from neighboring blocks, non-target text regions are masked before recognition. We rasterize line boxes belonging to the current block into \(M_t^{\mathrm{own}}\), and line boxes from other blocks overlapping the crop into \(M_t^{\mathrm{nt}}\). The masked region is
\begin{equation}
M_t = M_t^{\mathrm{nt}} \setminus M_t^{\mathrm{own}}.
\end{equation}
The masked pixels are filled with a local background color estimated from non-text pixels in the crop. If no reliable non-text pixels are available, the local crop appearance is used as the fallback background estimate.

Finally, the processed crop is placed at the center of a square canvas:
\begin{equation}
s_t = \alpha \cdot \max(w_t,h_t),
\end{equation}
where \(w_t\) and \(h_t\) are the width and height of the processed crop, \(s_t\) is the side length of the square canvas, and \(\alpha\) is a fixed padding scale. The remaining canvas area is filled with the same local background color. This padding strategy regularizes the input shape while preserving the original crop aspect ratio.

\end{document}